\documentclass[review,onefignum,onetabnum]{siamart171218}

\usepackage{lipsum}
\usepackage{amsfonts}
\usepackage{graphicx}
\usepackage{epstopdf}
\usepackage{algorithmic}
\usepackage{multirow}
\usepackage{float}
\ifpdf
  \DeclareGraphicsExtensions{.eps,.pdf,.png,.jpg}
\else
  \DeclareGraphicsExtensions{.eps}
\fi

\newcommand{\BigO}[1]{\ensuremath{\operatorname{O}\bigl(#1\bigr)}}
\renewcommand{\algorithmiccomment}[1]{\bgroup\hfill//~#1\egroup}

\newsiamremark{remark}{Remark}
\newsiamremark{hypothesis}{Hypothesis}
\crefname{hypothesis}{Hypothesis}{Hypotheses}
\newsiamthm{claim}{Claim}

\headers{GRAPH EMBEDDING METHODS}{M. Xu}

\title{UNDERSTANDING GRAPH EMBEDDING METHODS AND THEIR APPLICATIONS\thanks{
\funding{This work was funded by the NIH grant U01 HL1163232 and a J-Clinic for Machine Learning in Health award at MIT.}}}

\author{Mengjia Xu
\thanks{McGovern Institute for Brain Research, Massachusetts Institute of Technology, Cambridge, MA\and Division of Applied Mathematics, Brown University, Providence, RI (\email{mengjia\_xu1@hotmail.com})}
}

\usepackage{amsopn}
\usepackage{amssymb}

\makeatletter
\newcommand*{\addFileDependency}[1]{% argument=file name and extension
  \typeout{(#1)}% latexmk will find this if $recorder=0 (however, in that case, it will ignore #1 if it is a .aux or .pdf file etc and it exists! if it doesn't exist, it will appear in the list of dependents regardless)
  \@addtofilelist{#1}% if you want it to appear in \listfiles, not really necessary and latexmk doesn't use this
  \IfFileExists{#1}{}{\typeout{No file #1.}}% latexmk will find this message if #1 doesn't exist (yet)
}
\makeatother

\newcommand*{\myexternaldocument}[1]{%
    \externaldocument{#1}%
    \addFileDependency{#1.tex}%
    \addFileDependency{#1.aux}%
}
\ifpdf
\hypersetup{
  pdftitle={UNDERSTANDING GRAPH EMBEDDING METHODS AND THEIR APPLICATIONS},
  pdfauthor={M. Xu}
}
\fi

\myexternaldocument{ex_supplement}

\begin{document}

\maketitle

\begin{abstract}
Graph analytics can lead to better quantitative understanding and control of complex networks but traditional methods 
suffer from high computational cost and excessive memory requirements associated with the high-dimensionality
and heterogeneous characteristics of industrial size networks. 
Graph embedding techniques can be effective in converting high-dimensional sparse graphs into low-dimensional, dense and continuous vector spaces, preserving maximally the graph structure properties.
Another type of emerging graph embedding employs Gaussian distribution-based graph embedding with important
uncertainty estimation.
The main goal of graph embedding methods is to pack every node's properties into a vector with a smaller dimension, hence, node similarity in the original complex irregular
spaces can be easily quantified in the embedded vector spaces using standard metrics.
The generated nonlinear and highly informative graph embeddings in the latent space can be conveniently used to address different downstream graph analytics tasks (e.g., node classification, link prediction, community detection, visualization, etc.). 
In this Review, we present some fundamental concepts in graph analytics and graph embedding methods, focusing 
in particular on random walk-based and neural network-based methods. We also discuss the emerging 
deep learning-based dynamic graph embedding methods. We highlight the distinct advantages of
graph embedding methods in four diverse applications, and present implementation details and references 
to open-source software as well as available databases in the Appendix for the interested readers to start their exploration
into graph analytics.
\end{abstract}



\section{Introduction}
Graphs are a universal language for describing and modeling complex systems. Network data are ubiquitous across diverse application fields, such as brain networks~\cite{rosenthal2018mapping,xu2019gaussian,xu2020graph} in brain imaging, molecular networks~\cite{liu2019chemi} in drug discovery, protein-protein interaction networks~\cite{kashyap2018protein} in genetics, social networks~\cite{wang2019user} in social media, bank-asset networks~\cite{zhou2019asset} in finance,  publication networks~\cite{west2016recommendation} in scientific collaborations, etc. 
% does alphaFold uses graphs? how? can you see from ther Github code?
Unlike the regular grid-like Euclidean space data (e.g., images, audio and text), the aforementioned network data are from irregular non-Euclidean domains. However, as a nonlinear data structure, a graph can be used as an effective tool to describe and model the complex structure of network data. Specifically, a graph $G(V,E)$ is 
typically composed of a set of vertices (or entities) denoted by $V$, and certain  edges (or relations) denoted by $E$ between different vertices. Modeling complex systems as graphs facilitates the characterization of very useful high-order geometric patterns for the networks, which has a great impact on improving the performance of different network data analysis tasks, e.g., gene network reconstruction, protein function prediction, human face recognition, etc.

Graph analytics (also known as network analysis) has become an exciting and impactful research area in recent years. Developing effective and efficient graph analytics can greatly help to better understand complex networks. However, traditional graph analytics including path analysis, connectivity analysis, community analysis and centrality analysis, are largely based on extracting handcrafted graph topological features directly from the adjacency matrices. When we apply these methods to large-scale network analysis in industrial systems, they may suffer from high computational cost and excessive memory requirements due to the challenging and inevitable high-dimensionality and emerging heterogeneous characteristics of the original networks~\cite{su2020network}. Moreover, the hand-engineered features are usually task-specific and cannot get equivalent performance while employing them for different tasks. 

\begin{figure}[htbp]
    \centering
    \includegraphics[width=.85\textwidth]{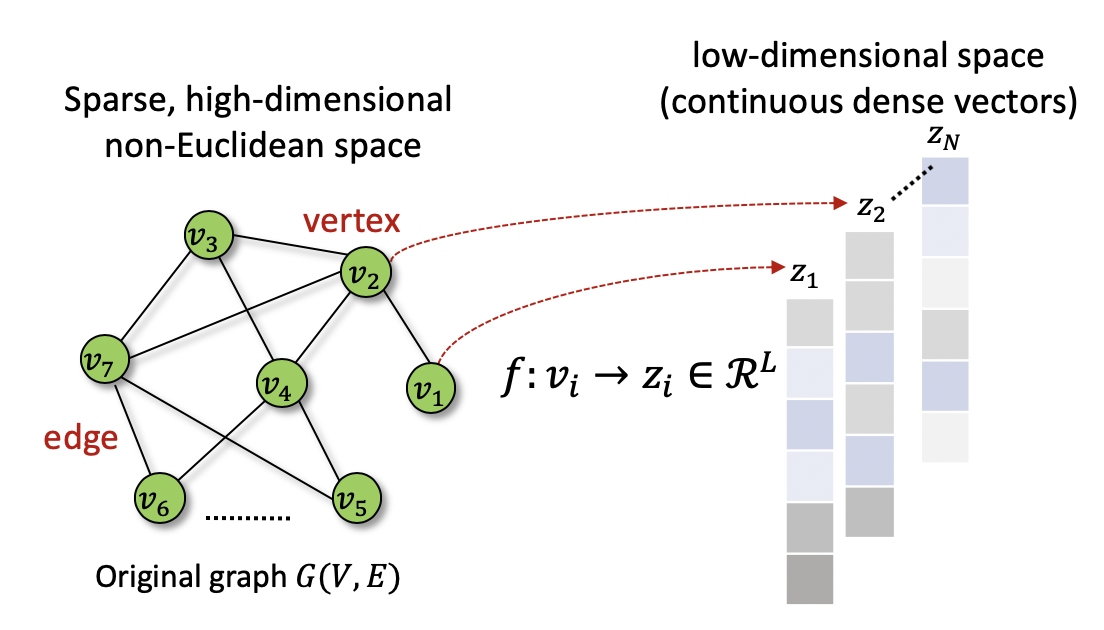}
    \caption{\textbf{Schematic of graph (node) embedding}. For a simple graph $G(V,E)$ consisting of a node set $V$ and an edge set $E$, using a graph embedding model $f$, different nodes (e.g., $v_1$ and $v_2$) from the original graph in a high-dimensional irregular domain can be mapped into a latent low-dimensional space as a $L$-dimensional dense and continuous vector $z_i$, $L\ll|V|$. The node structure property can be also preserved in the latent space, i.e., similar nodes in the original space will be close to each other in the latent space. Moreover, the obtained latent variables $z_i,i\in V$ (i.e., features) can be readily used for diverse downstream graph analytic tasks.}
    \label{fig:fig1}
%%%%%%%%%%%%%%%%%%%%%%%%%%%%%%%%%%%%%%%%%%%%
\end{figure}

Recently, graph embedding techniques have shown remarkable capacity of converting high-dimensional sparse graphs into low-dimensional, dense and continuous vector spaces (see \Cref{fig:fig1}), where graph structure properties are maximally preserved ~\cite{cai2018comprehensive}. The generated nonlinear and highly-informative graph embeddings (or features) in the latent space can be conveniently used to address different downstream graph analytic tasks (e.g., node classification, link prediction, community detection, visualization, etc.). The main aim of graph embedding methods is to encode nodes into a latent vector space, i.e., pack every node's properties into a vector with a smaller dimension. Hence, node similarity in the original complex irregular spaces can be easily quantified based on various similarity measures (e.g., dot product and cosine distance) in the embedded vector spaces. Furthermore, the learned latent embeddings can greatly support much faster and more accurate graph analytics as opposed to directly performing such tasks in the high-dimensional complex graph domain.

%%%%%%%%%%%%%%%%%%%%%%%%%%%%%%%%%%%%%%%%%%%%

A broad taxonomy of graph embedding methods used in the literature points to three major categories~\cite{cui2018survey}: matrix factorization-based methods, random walk-based methods, and neural network-based methods. Matrix factorization-based methods (e.g.,~\cite{cao2015grarep,belkin2002laplacian,ou2016asymmetric,roweis2000nonlinear}) construct a high-order proximity matrix based on transition probabilities and factorize it to obtain the node embeddings, but they cannot easily scale up to large network embeddings. In this paper, we focus on the more scalable random walk-based methods and neural network-based methods. Furthermore, in addition to the static graph embedding methods, we also discuss the emerging deep learning-based {\em dynamic} graph embedding methods. Finally, we highlight the distinct advantages of graph embedding methods in four diverse applications, and conclude with a summary. In the supplementary material, we include information about open-source software, available data and implementation details.

\section{Mathematical Formulation of the Graph Embedding Problem}
\label{sec:section_2}
In this section, we first present some preliminary graph notations, definitions and properties used in graph embedding. Then, we review the node similarity measures frequently applied in graph embedding. Finally, we formulate the specific graph embedding problem setting.

\subsection{Preliminaries}
We consider a graph $G(V,E)$ as a mathematical data structure that contains a node (or vertex) set $V=\{v_1, v_2, v_3, ..., v_n\}$ and an edge (or link) set $E$. In the edge set, one edge $e_{ij}$ describes the connection between two different nodes $v_i$ and $v_j$, hence, $e_{ij}$ can be represented as $(v_i, v_j)$, where $v_i, v_j\in V$, and nodes $v_i$ and $v_j$ are adjacent nodes. 
From different perspectives (i.e., edge direction, diversity of nodes and edges, edge cost), graphs can be classified into different categories: directed/undirected graphs, homogeneous/heterogeneous graphs or weighted/binary graphs. We present more details below.

\begin{figure}[htbp]
    \centering
    \includegraphics[width=.9\textwidth]{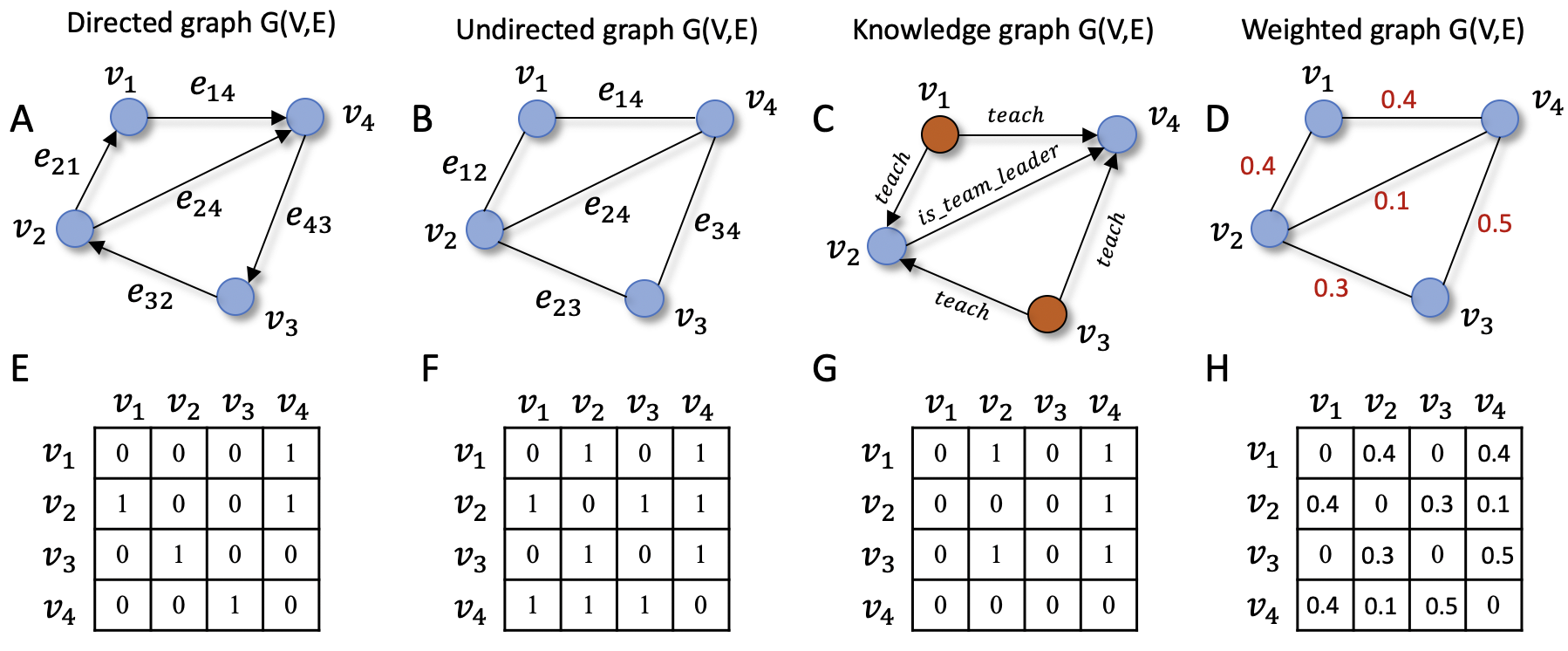}
    \caption{\textbf{Different types of graphs and their corresponding adjacency matrix representations}. The first row from (A) to (D) are, respectively, directed, undirected, knowledge and weighted graph examples. The main difference between (A) and (B) is that edges are directed in (A) but undirected in (B). (C) is a knowledge graph consisting of two different types of nodes (in ``brown'' and ``blue'' colors) and two different types of edges (``teach'' and ``is\_team\_leader''). Graph (C) is an instance of directed and heterogeneous graph. (D) shows a weighted graph where every edge is weighted with a specific value. The second row from (E) to (H) shows the corresponding $4 \times 4$ adjacency matrices for graphs (A)-(D).}
    \label{fig:graph_types}
\end{figure}

\textbf{1) Directed/undirected graphs}: Regarding the ``directed graph'' (also called digraph) as shown in \Cref{fig:graph_types}(A), every edge has a specific direction, e.g., flight networks, Google maps. Conversely, in an ``undirected graph'', edges have no directions, see an example in \Cref{fig:graph_types}(B). Since edges in undirected graphs are un-ordered pairs, the edge $e_{12}$ can be also replaced with $e_{21}$. An undirected graph can be also viewed as a bidirectional graph. 

\textbf{2) Homogeneous/heterogeneous graphs}: For homogeneous graphs, all nodes or edges are of the same type, e.g., friendship network with nodes denoting different persons and edges denoting their friendship. In contrast, heterogeneous graphs consist of multiple types of nodes and/or edges. A knowledge graph is a typical directed heterogeneous graph, see an example in \Cref{fig:graph_types}(C) showing an education network. Different node colors refer to different node types, i.e., brown color nodes 
stand for ``Teacher'' type while blue ones stand for ``Student'' type. Moreover, there are two edge types (``teach'' and ``is\_team\_leader'') in the knowledge graph.
%
%%%%%%%%%%%%%%%%%%%%%%%%%%%%%%%%%%%%%%%%%
%%%%%%%%%%%%%%%%%%%%%%%%%%%%%%%%%%%%%%%%%

\textbf{3) Weighted/binary graphs:} For weighted graphs, every edge is assigned a specific numerical value (i.e., weight), see an example in \Cref{fig:graph_types}(D), whereas binary graphs (or unweighted graphs) do not have any weight associated with edges.

\begin{figure}[htbp]
    \centering
    \includegraphics[width=.93\textwidth]{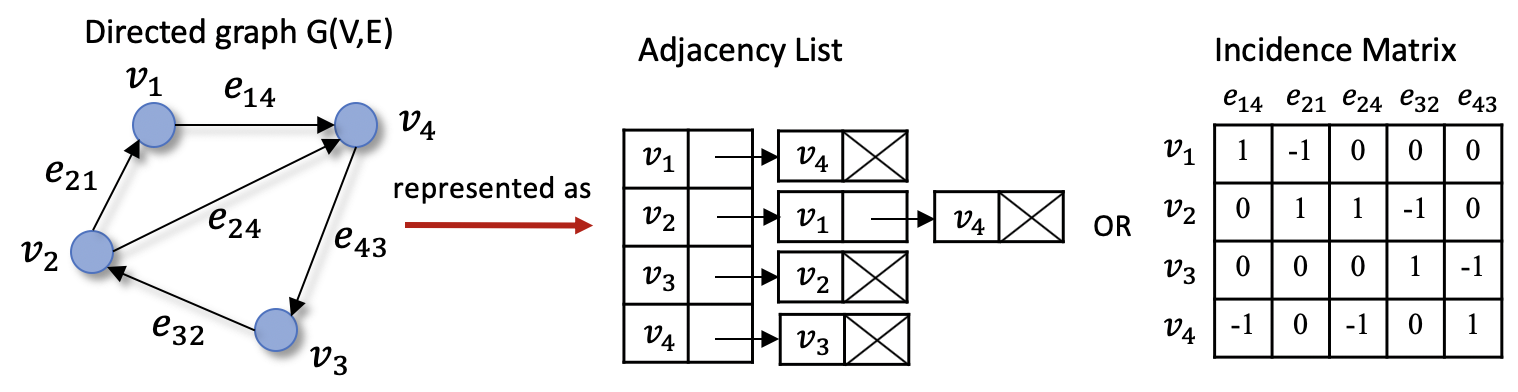}
    \caption{\textbf{Different ways to represent a static graph.} Beyond the adjacency matrix representation as shown in \Cref{fig:graph_types} (E-H), a static graph can be also represented as an ``adjacency list'' or ``incidence matrix''.}
    \label{fig:fig3}
\end{figure}

According to different graph edge density and types of operations applied on a graph, a graph can be represented in three different ways as: 
\textbf{i) Adjacency matrix ($A$):} $A$ is a $|V|\times |V|$ 2D square matrix with its element $\{s_{i,j}| i, j \in (0,|V|]\}$ representing whether two nodes are connected in a graph, and $|V|$ is the total number of vertices in the graph. Specifically, if nodes $v_i$ and $v_j$ are connected, $s_{i,j} = 1$ for a binary graph (note that, $s_{i,j} = w$ for a weighted graph, $w$ is the edge weight), otherwise, $s_{i,j} = 0$. Moreover, $A$ is a symmetric matrix for undirected graphs and asymmetric for the directed graphs, see examples in \Cref{fig:graph_types}(E-H). Furthermore, the adjacency matrix has a high computational cost for a traversal of every node's adjacent nodes of the order of $O(|V|)$, and therefore, it is not  appropriate for representing large sparse  graphs due to high space cost $O(|V|^2)$.  
\textbf{ii) Adjacency list:} It makes use of a linked list to only store each node's adjacent nodes (see a detailed example in \Cref{fig:fig3}). Therefore, the adjacency list is convenient for node operations (i.e., insert, delete or add nodes), and the space cost is only $O(|E|)$, which benefits effective representations for large sparse graphs (i.e., $|E|\ll|V|\times|V|$). 
\textbf{iii) Incidence matrix:} It employs a $|V|\times |E|$ matrix to represent the relationship between node set and edge set in a graph. More details can be seen in \Cref{fig:fig3}, where each column denotes different edges, and each row denotes different nodes in a directed graph. Each element in the matrix can be filled with (``1"- the column edge is one outgoing edge from the row node; ``0''- the column edge is not connected with the row node; ``-1''- the column edge is one incoming edge to the row node (for undirected graphs, elements with ``-1'' are all filled with ``'1').
%
%%%%%%%%%%%%%%%%%%%%%%%%%%%%%%%%%%%%%%%%%%%%%
%%%%%%%%%%%%%%%%%%%%%%%%%%%%%%%%%%%%%%%%%%%%%

\subsection{Graph embedding problem setting}
\label{sec_2.2}
In line with the aforementioned graph notations and definitions, given a graph $G=(V,E)$, the task to learn its graph node embeddings (e.g., $L$ dimension, $L\ll|V|$) can be mathematically formulated as learning a projection $\phi$, such that all graph nodes ($V = \{v_i|i=1,2,...,|V|\}$) can be encoded as two different embedding forms from a high-dimensional space into a low-dimensional space. One node embedding form is deterministic point vectors ($\Phi = \{z_i \in \mathbb{R}^L | i=1,2,...,|V|\}$) while another one is stochastic probabilistic distributions ($\Phi = \{\mathbb{P}_i \sim \mathcal{N}(\mu_i,\Sigma_i)|i=1,2,...,|V|\}$), where the mean vector $\mu_i \in \mathbb{R}^{L/2}$, and the covariance matrix $\Sigma_i \in \mathbb{R}^{L/2\times L/2}$. Moreover, graph structure properties are maximally preserved in the embedded space. To this end, node pairwise similarity (e.g., dot product-$z_j^Tz_i$) in the embedded latent space facilitates an approximation of the corresponding node similarity ($Sim(v_i, v_j)$) in the original space, i.e., $Sim(v_i, v_j) \approx z_j^Tz_i$ for vector-based graph embedding;
specifically, $Sim(v_i, v_j) \approx \mu_j^T\mu_i$ for Gaussian distribution-based graph embedding. Here, $Sim$ is a pre-defined similarity function (see \cref{tab:similarity_embedding_space} in the Appendix and more details in \cref{sec_sim_latent_space}). The specific graph node embedding formula is shown in \cref{eq:mapping},
i.e., 
\begin{equation}
\phi:v_i \rightarrow \left\{
\begin{array}{l}
z_i \in \mathbb{R}^L \;\; (i = 1,2,..., |V|)  \; \; or\\
P_i \sim \mathcal{N}(\mu_i,\Sigma_i), \mu_i \in \mathbb{R}^{L/2}, \Sigma_i \in \mathbb{R}^{L/2\times L/2} \;\; (i = 1,2,..., |V|).
\end{array}
\right.
\label{eq:mapping}
\end{equation}

Generally, the main goal of graph embedding is to encode nodes into low-dim ensional space, such that similarity in the latent embedded space approximates similarity in the original high-dimensional graph, while preserving the graph structure property. Essentially, most of the advanced graph node embedding techniques learn low-dimensional node representations by solving an optimization problem, which follows an {\em unsupervised} learning schema and are independent of downstream prediction tasks. Specifically, graph embedding approaches contain three major components: graph structure property preservation, node similarity measurement in both the original and latent space, and an encoder. Next, we present more details about these three key components.

\subsubsection{Graph structure property preservation}
\label{sec_2.2.1} 
Proximity measures play an important role in the quantification of graph structure property preservation for diverse graph embedding methods. Specifically, we present three different types of proximity measures: 

1) \textit{First-order proximity}: It is used to characterize local pairwise similarity between nodes linked by edges, i.e., local network structure property. However, it is insufficient for preserving the global network structure. First-order proximity between two nodes is equal to their edge weight  ($w_{ij} = 1$ for a binary graph). Moreover, in order to develop a graph embedding model preserving the first-order proximity, the objective function ($\mathbb{E}$) can be constructed as in \cref{eq:1st-order-obj}, where $\hat{p}_1(v_i,v_j)$ and $p_1(v_i,v_j)$ represent the corresponding empirical distribution and model distribution of first-order proximity, respectively, which are modeled as joint probabilities.
\begin{equation}
\begin{array}{ll}
\mathbb{E} = D_{KL}(\hat{p_1}||p_1)=-\sum\limits_{i,j \in E}w_{ij}\log(p_1(v_i,v_j)), \\
\text{where} \;\; \hat{p}_1(v_i,v_j) = \frac{w_{ij}}{\sum\limits_{s,t \in E}w_{st}}, \;\; p_1(v_i,v_j) = \frac{exp(z_i^{T} z_j)}{\sum\limits_{s,t \in E} exp(z_s^{T} z_t)}.
\end{array}
\label{eq:1st-order-obj}
\end{equation}
Here, $v_{i}$ and $v_{j}$ represent the original two nodes $i$ and $j$; $z_{i}, z_{j}$ are the corresponding embedding vectors of $v_{i}$ and $v_{j}$; $w_{ij}$ denotes the weight between two nodes $i$ and $j$.

2) \textit{Second-order proximity}: It is used to capture the proximity between the neighborhood structures of the nodes~\cite{cai2018comprehensive}, i.e., the global network structure property. Second-order proximity can be easily estimated using the transitional probability metric between two nodes. In order to develop a graph embedding model preserving the second-order proximity, the objective function ($\mathbb{E}^{'}$) is defined as in \cref{eq:2nd-order-obj}, where $\hat{p}_2(v_i|v_j)$ and $p_2(v_i|v_j)$ represent the corresponding empirical distribution and model distribution of the neighborhood structure that are modeled as two different conditional probabilities, as follows:  
\begin{equation}
\begin{array}{ll}
\mathbb{E}^{'} = \sum\limits_{i\in V}{D_{KL}(\hat{p}_2(\cdot|v_i)||p_2(\cdot|v_i))}=-\sum\limits_{i,j \in E}w_{ij}\log(p_2(v_j|v_i)) \\
where \;\; \hat{p}_2(v_j|v_i) = \frac{w_{ij}}{\sum\limits_{k \in V}w_{ik}}, \;\; p_2(v_j|v_i) = \frac{exp(z_j^{{\prime}T} z_i)}{\sum\limits_{k \in V} exp(z_k^{{\prime}T} z_i)}.
\end{array}
\label{eq:2nd-order-obj}
\end{equation}
Here, $z_{i}^{\prime}$ is the representation of $v_i$ when it is treated as a specific “context” \cite{tang2015line}.

3) \textit{High-order proximity}: There are not many works in the literature focusing on high-order proximity quantification so far, as second-order proximity works well in most graph embedding methods. Usually, high-order proximity can be defined based on some metrics, such as Rooted PageRank~\cite{brin2017anatomy}. 

\subsubsection{Node similarity measures in the original graphs} 
\label{sec_2.2.2} 
Selecting a proper node similarity measure is particularly important for finding effective node context (or neighbors) aiding for optimizing diverse graph embedding models. Similarity between nodes in the original graphs can be characterized based on five different aspects: 1) presence of edge, 2) overlapped neighborhood, 3) reachable by k-hops, 4) reachable through random walks, and 5) similar node attributes. Accordingly, node similarity definitions commonly used in existing graph embedding methods mainly consist of three different types: multi-hop neighborhood-based~\cite{bojchevski2017deep}, random walk-based~\cite{perozzi2014deepwalk,grover2016node2vec,tang2015line}, and adjacency matrix-based methods~\cite{roweis2000nonlinear, belkin2002laplacian, ou2016asymmetric}. The key idea of graph embedding is to optimize low-dimensional embeddings to approximate the node similarities in the original graph generated by the above measures. 

\begin{figure}[htbp]
    \centering
    \includegraphics[width=.7\textwidth]{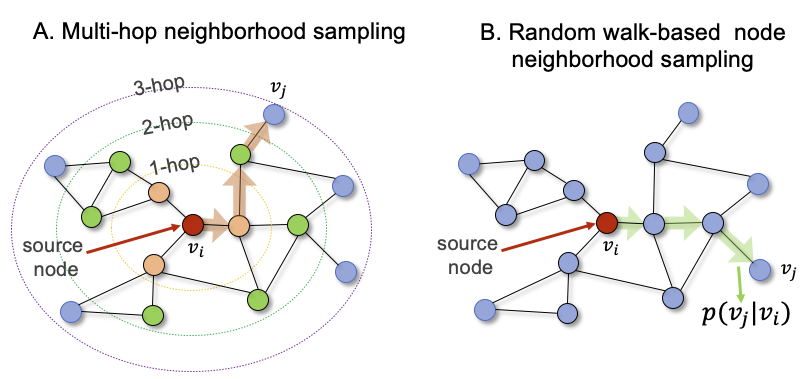}
    \caption{\textbf{Node neighborhood sampling techniques.} A) Multi-hop neighborhood-based node neighbor sampling. The red node ($v_i$) is taken as the source target node, and nodes of different colors represent the sampled nodes at different hops; $v_j$ is one of the 3-hop neighbors of $v_i$. The orange arrows mark the shortest path ranging from $v_i$ to $v_j$. B) Random walk-based node neighbor sampling strategy, every node neighborhood is sampled based on the transition probability computed using \cref{eq:rw}.}
    \label{fig:randomwalk}
\end{figure}

\textit{1) Multi-hop neighborhood-based node similarity}: The ``hop'' terminology originally appeared in the wired computer networking field, and ``hop count'' can provide a measure of distance between source host and destination host~\cite{kurose1992computing}. Recently, in order to sample node neighbors in the original graphs and preserve graph structure property, the multi-hop neighborhood (or $k$-hop neighborhood) method has become an effective solution to address this problem in graph embedding. Specifically, \textit{k}-hop neighborhood is defined as the set of vertices that are reachable from a source node in \textit{k} hops or fewer (i.e., following a path with \textit{k} edges or fewer in binary graphs), see an example of 3-hop neighborhood in \Cref{fig:randomwalk}(A). In particular, the \textit{k}-hop neighborhood searching method enables us to sample similar neighbors for every sampled source node at different hops in the original graphs. In addition, higher-order proximity can be preserved by increasing the number of hops. For instance,the G2G method~\cite{bojchevski2017deep} that we study in \cref{sec_3.2_gaussian_embedding}, evaluates the 2-hop and 3-hop node neighborhood sampling strategies for binary graph embedding, and tests show that $k = 2$ is sufficient for preserving graph structure property in high-order proximity.
%%%%%%%%%%%%%%%%%%%%%%%%
%%%%%%%%%%%%%%%%%%%%%%%%

\textit{2) Random walk-based node similarity}: Random walks have been used as stochastic similarity measures for various problems (e.g., content recommendation~\cite{nikolakopoulos2019recwalk}, community detection~\cite{okuda2019community}, image segmentation~\cite{bertasius2017convolutional,xu2017deep}). In particular, the node similarity ($Sim(v_i, v_j)$) in the complex graph can be defined as the probability $p(v_j|v_i)$ of reaching a node $v_j$ on a random walk ($W_{v_i} = {r_0, r_1, ..., r_l}$) of length $l$ over the graph from the source node $r_0 = v_i$, see a random walk example in \Cref{fig:randomwalk}(B). Particularly, the nodes in the random walk are generated based on the distribution described in \cref{eq:rw}, where $r_i$ denotes the $i-th$ node in the random walk starting from $r_0 = u$, $\pi_{vx}$ denotes un-normalized transition probability between nodes $v$ and $x$, and $Z$ is a normalizing constant~\cite{grover2016node2vec}. 
%%%%%%%%%%%%%%%%%%%%%%%%%%%
\begin{equation}
p(r_i = x|r_{i-1} = v) = \left\{
\begin{aligned}
\pi_{vx}/Z & &if (v,x) \in E, \\
0 & &otherwise.
\end{aligned}
\right.
\label{eq:rw}
\end{equation}
%%%%%%%%%%%%%%%%%%%%%%%%%%%

In the earlier DeepWalk methods~\cite{perozzi2014deepwalk}, a random walk is generated (similar as a sampled``sentence'' in nature language processing applications) for each node (comparable as ``word'' in the language modeling problems) in the graph following the random search strategy as shown in \cref{eq:rw}. It then learns the node embeddings through predicting the nearby nodes co-occured on the generated random walk using a language embedding model (i.e., SkipGram model~\cite{mikolov2013efficient}). Moreover, other methods that we discuss in \cref{sec:section_3.1}, such as LINE~\cite{tang2015line} and node2vec~\cite{grover2016node2vec}, also apply modified random walk strategies to sample similar neighbors (or context) for different nodes in the original graph. Compared with the  multi-hop neighborhood-based measure, the major advantage of the random walk-based similarity measure is that it does not need to train all the node pairs but it only considers the node pairs that co-occur on random walks.

\textit{3) Adjacency matrix-based node similarity}: Node similarity can be also characterized by the presence of edges between nodes in the adjacency matrices of original graphs. For example, the classic matrix factorization-based graph embedding methods  (e.g., Locally linear embedding ~\cite{roweis2000nonlinear}, Laplacian eigenmaps~\cite{belkin2002laplacian}, HOPE~\cite{ou2016asymmetric}) (see \cref{sec:Applications}) and stochastic neighborhood embedding (SNE) method~\cite{hinton2003stochastic} all employ adjacency matrix-based node similarity measures for graph embedding. They have achieved good performance in graph reconstruction problems. However, using the adjacency matrix-based node similarity measure to carry out graph embedding, the obtained node embeddings usually overfit the adjacency matrix of the original graph. Hence, it does not work well on inconspicous connection detection (such as downstream link prediction tasks)~\cite{nerurkar2019survey}. Moreover, due to the computational complexity $O(|V|^2)$, using adjacency matrix-based measures is hard to scale up to large-scale networks. Beyond these limitations, adjacency matrix-based similarity measures only consider local connections. 

\subsubsection{Similarity measures in the embedded space}
\label{sec_sim_latent_space}
According to the specific learned node representation/embedding form (e.g., vector point representations, Gaussian distribution representations) in the latent space using different graph embedding methods, we can employ different metrics ($M$) to measure the node similarity in the low-dimensional embedded space. Specifically, given two randomly selected nodes $v_i$, $v_j$ from an original graph $G$, where $z_i, z_j$ are the corresponding point vector node embeddings in the latent space, the similarity measures ($M$) for the point vector node embeddings can be computed by measures, such as dot product, cosine similarity and Euclidean distance (see detailed formulas in \cref{tab:similarity_embedding_space} of \cref{sec:similarity_latent_space} in the Appendix). However, for node embeddings as Gaussian distributions in the latent space, the similarity measures frequently used in recent works involve expected likelihood (EL), KL-divergence ($D_{KL}$), and the Wasserstein distance ($W_2$) (see  \cref{tab:similarity_embedding_space} in the Appendix). $
P_i \sim \mathcal{N}(\mu_i,\Sigma_i)$ and $P_j \sim \mathcal{N}(\mu_j,\Sigma_j)$ are the learned stochastic node representations (i.e., multivariate Gaussian distributions) in the latent space for nodes $v_i$ and $v_j$, where $\mu_i$ and $\Sigma_i$ denote the learned mean vector and covariance matrix for node $v_i$, and $\mu_j$ and $\Sigma_j$ are the learned mean vector and covariance matrix for node $v_j$.

\section{Overview of graph embedding methods}
\label{sec:section_3}
According to the aforementioned mathematical problem settings of graph embedding in \cref{sec:section_2}, we further subdivide the prevalent graph embedding methods into three main categories: 1) vector point-based graph embedding, 2) Gaussian distribution-based graph embedding, and 3) dynamic graph embedding. More details about these three types of methods are presented below.

\subsection{Vector point-based graph embedding methods}
\label{sec:section_3.1}
Vector point-based graph embedding methods can be further divided into three main types: a) matrix factorization-based methods(GraRep~\cite{cao2015grarep}, HOPE~\cite{ou2016asymmetric}), b) random walk-based methods (e.g., DeepWalk~\cite{perozzi2014deepwalk}, LINE~\cite{tang2015line}, node2vec~\cite{grover2016node2vec}), and c) deep learning-based methods (SDNE~\cite{wang2016structural}). The main purpose of vector point-based graph embedding is to project high-dimensional graph nodes into low-dimensional vectors in a latent space, while preserving the original graph structure properties, see a specific definition in \cref{eq:mapping}. Nodes that are ``close'' in the original graph are embedded to a latent space with similar ``vector representations''. The ``closeness'' between variant nodes in the original space can be modeled in different ways by using the measures presented in the \cref{sec_2.2.2}. Specifically, matrix factorization-based graph embedding methods measure the ``closeness'' degree between nodes via the adjacency matrix-based measures, while random walk-based graph embedding methods apply random walk techniques to extract close node neighbors. 

\begin{figure}[htbp]
    \centering
    \includegraphics[width=.96\textwidth]{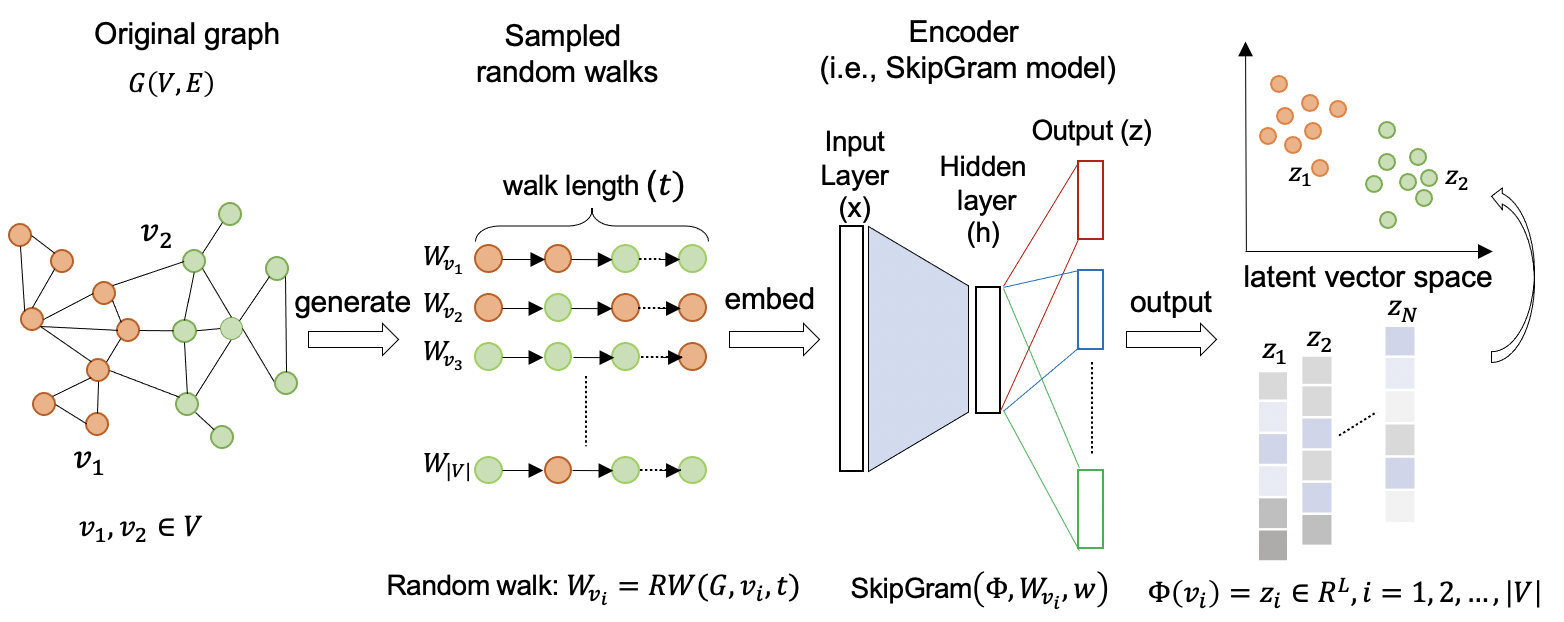}
    \caption{Pipeline for random walk-based graph embedding methods. In order to learn node embeddings for the original graph $G=(V,E)$ consisting of two types of nodes marked in different colors, we apply random walks methods to first generate a set of node context ($W_{v_i}$) for every node ($v_i \in V$); the sampled node context (i.e., random walks) are of same fixed walk length $t$. Second, based on the generated node contexts, a language embedding model playing the role of an encoder such that every node is represented as a low-dimensional, continuous vector in the latent space. The distance (e.g., dot product, cosine similarity or Euclidean distance) between vectors (or ``node embeddings'') in the latent vector space approximates the similarity in the original graph. Additionally, the learned vectors can be simply mapped to 2D space as points using dimension reduction techniques (e.g., t-SNE, MDS, PCA). The learned node embedding features ($\Phi \in \mathbb{R}^{|V|\times L}$) for all nodes can be readily and efficiently used for different downstream tasks, such as link prediction, node classification, community detection, etc.}    
    \label{fig:vec_emb_workflow}
\end{figure}

Since we have reviewed the advantages and limitations of adjacent matrix-based methods in 3) of \cref{sec_2.2.2}, here, we mainly focus on summarizing the random walk-based graph embedding methods including the main pipeline and key techniques. Given a graph $G(V,E)$, the main pipeline of random walk-based graph node embedding method for $G$ is shown in \Cref{fig:vec_emb_workflow}. It mainly contains three key stages, which are described in detail, as follows.
%%%%%%%%%%%%%%%%%%%%%%%%%%%%%%%%%%%%%%%%%%%%%%%%%%
%%%%%%%%%%%%%%%%%%%%%%%%%%%%%%%%%%%%%%%%%%%%%%%%%%

\begin{figure}
    \centering
    \includegraphics[width=.5\textwidth]{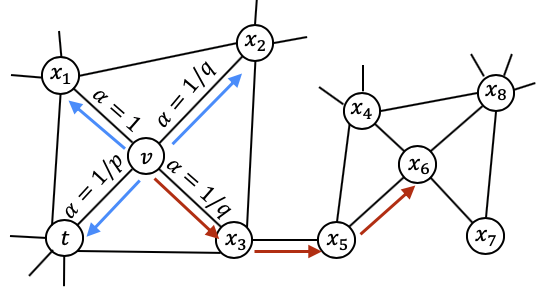}
    \caption{\textbf{Biased random walk schema incorporating both BFS and DFS.} Starting from the source target node ($v$) to generate a random walk with fixed length ($l = 4$) over the graph, blue arrows show the breadth-first search (BFS) directions while red arrows indicate the depth-first search (DFS) directions. The DFS strategy helps expanding the node neighbors deeper and enables exploration of node neighbors (e.g., $x_6$) that have similar structure roles as the starting node (i.e., $v$) beyond the directly connected node neighbors (e.g., $x_1, x_2, x_3$). Hence, 2nd-order proximity can be also preserved. The biased random walk strategy proposed in node2vec~\cite{grover2016node2vec} achieves a trade-off of BFS and DFS; the explored neighborhood search is guided by using a search bias parameter ($\alpha$), which is defined by the return rate ($p$) and exploration rate ($q$) as shown in \cref{eq:biased_rw}. We provide more details about these two parameters in 1) in \cref{sec:section_3.1}.}
    \label{fig:biased_rw}
\end{figure}

\textbf{1) Node context generation based on random walks.} Motivated by the ``word2vec'' method~\cite{mikolov2013efficient} developed for learning word representations based on sentences, the authors of DeepWalk~\cite{perozzi2014deepwalk} first adopted the random walk-based node neighbors sampling technique (see \cref{sec_2.2.2}) to extract node context (similar as ``sentence''), and then employed an encoder called ``SkipGram'' model~\cite{mikolov2013efficient} to learn graph node embeddings based on the sampled similar node pairs in the random walks. In this way, it can effectively encode the structure and topological information from the original graph into the latent space. However, DeepWalk~\cite{perozzi2014deepwalk} can only capture the local structure information by using the truncated random walks, but the global structure information is missing. To address this problem, node2vec~\cite{grover2016node2vec} employs an improved {\em biased random walk} method to sample node context by considering both local and global structure information from the original graph. \Cref{fig:biased_rw} shows a detailed schematic of node neighbor expansion using the biased random walk approach (or ``2nd-order random walk''), which incorporates both BFS (breadth-first-search) and DFS (depth-first-search) searching strategies with two ratio parameters ($p$ and $q$). In general, nodes in a random walk are generated by using the formula defined in \cref{eq:rw}, however, in the ``biased random walk'', the unnormalized transition probability $\pi_{vx}$ in \cref{eq:rw} is modified using a search bias ($\alpha$) in conjunction with edge weight $k_{vx}$ ($k_{vx} = 1$ if binary graphs) to guide node neighbor searching. Specifically, assume a random walk just traversed nodes $t, v$, and resides at $v$, then the unnormalized transition probability ($\pi_{vx}$) between the node $v$ and the next walk node $x$ can be computed by the formula in \cref{eq:biased_rw} as follows. 
%%%%%%%%%%%%%%%%%%%%%%%%%%%%%
\begin{equation}
\pi_{vx} = \alpha_{pq}(t,x)\cdot k_{vx}, \;\; where \;\;
\alpha_{pq}(t,x) = \left\{
\begin{aligned}
1/p & & if &&d_{tx} = 0,\\
1   & & if &&d_{tx} = 1,\\
1/q & & if &&d_{tx} = 2,\\
\end{aligned}
\right.
\label{eq:biased_rw}
\end{equation}
%%%%%%%%%%%%%%%%%%%%%%%%%%%%%
where the search bias ($\alpha_{pq}$) is defined by a return rate parameter ($p$) and an ``in-out'' exploration rate parameter ($q$); $d_{tx}$ represents the shortest distance between the previous visited node $t$ and the next visiting nodes $x$; $p$ and $q$ enables to control how fast the walk explores and leaves the starting node ($v$) (see \Cref{fig:biased_rw}). Particularly, the larger is $p$ ($> max(q,1)$), the less likely is to sample an already visited node, i.e., {\em moderate exploration}; inversely, when $p$ ($< min(q,1)$), it will lead to backtrack the step, which results in {\em local neighborhood sampling}. Moreover, $q$ is used to control the exploration of ``inward'' and ``outward'', i.e., when $q > 1$, it approximates the BFS behavior, and leads the random walks towards a {\em micro-view} of the node neighborhoods. Whereas, $q < 1$ is ``DFS-like'' {\em macro-view exploration} for the node neighborhoods. Additionally, when $p = q$, the node2vec method is the same as DeepWalk.
%%%%%%%%%%%%%%%%%%%%%%%%%%%%%
%%%%%%%%%%%%%%%%%%%%%%%%%%%%%

Moreover, the selection of window size ($w$) and walk length ($t$) also plays an important role in generating effective node context, because it can implicitly influence the number of node neighbors covered in the random walks, and affects the constraints of node similarity. The larger walk length $t$ is used, the more noisy node co-occurances would be introduced in the graph embedding. 

\textbf{2) Learn node embedding with an encoder.} Aiming to transform the graph nodes to low-dimensional vectors, the encoder is a key and necessary component in graph embedding techniques. The ``SkipGram model''~\cite{mikolov2013efficient} is a typical language embedding model, which can be also employed as an effective encoder for prevalent graph node embedding problems by feeding the generated node context from the step 1) and output the low dimensional node embeddings. Specifically, learning a graph node embedding ($\Phi =\{z_i\}_{i=1}^{|V|}, z_i \in \mathbb{R}^L$) corresponds to minimizing the negative log-likelihood of observing its neighborhood nodes ($v \in N_R(u)$) conditioned on the predicted source node's embedding ($z_u$), which is the output of the encoder (i.e., the SkipGram model). The parameters of the embedding model can be learned by minimizing the cross-entropy loss function ($\mathcal{L}$) as shown in \cref{eq:objfunc_rwEmb}.
%%%%%%%%%%%%%%%%%%%%%%%%%%%%%%%%%%%%%%%%
\begin{equation}
\mathcal{L} = -\log \sum_{u \in V}\sum_{v \in N_R(u)} p(v|z_u) = -\log \sum_{u \in V}\sum_{v \in N_R(u)} \frac{exp(z_u^Tz_v)}{\sum\limits_{n \in V}{exp(z_u^Tz_n)}}, 
\label{eq:objfunc_rwEmb}
\end{equation}
%%%%%%%%%%%%%%%%%%%%%%%%%%%%%%%%%%%%%%%%
where $V$ refers to the entire node set of the original graph, $N_R(u)$ denotes the sampled neighbor set for the source node $u$ using some neighborhood sampling strategy $R$ (here, we refer to ``random walks''), and $v$ refers to the nearby nodes of node $u$. In practice, DeepWalk~\cite{perozzi2014deepwalk} and node2vec~\cite{grover2016node2vec} have the same objective function as shown in the first part of \cref{eq:objfunc_rwEmb}, but they adopt different optimization policies. In paticular, DeepWalk~\cite{perozzi2014deepwalk} is optimized based on the hierarchical softmax~\cite{perozzi2014deepwalk}; the simplified loss function with the hierarchical softmax is shown in the last part of \cref{eq:objfunc_rwEmb}), which we need to normalize with respect to all nodes with high computational complexity ($O(|V)^2$). However, node2vec~\cite{grover2016node2vec} applies stochastic gradient descent (SGD) with ``negative sampling'' for more efficient model optimization. According to \cref{eq:objfunc_rwEmb}, the simplified objective function ($\mathcal{L}^{'}$) for node2vec using ``negative sampling'' is shown in \cref{eq:objfun_neg_sampling} as follows:
%%%%%%%%%%%%%%%%%%%%%%%%
\begin{equation}
\mathcal{L}^{'} = -\sum_{u \in V}\sum_{v \in N_R(u)}(\log(\sigma(z_u^Tz_v))-\sum_{i=1}^{k}\log(\sigma(z_u^Tz_{n_i}))), n_i \sim P_v,
\label{eq:objfun_neg_sampling}
\end{equation}
%%%%%%%%%%%%%%%%%%%%%%%%%%%%%%%%%%%%%%%%
where $\sigma$ denotes the sigmoid function; $n_i$ denotes the random sampled negative node neighbors, which follows a random distribution over all nodes. To learn graph node embeddings with the objective function in \cref{eq:objfun_neg_sampling}, we just need to normalize against ``$k$'' random negative samples, which can effectively speed up the training process and greatly reduce the computational complexity.

\textbf{\textbf{3) Output low-dimensional node embeddings for downstream tasks.}} With the node embedding learning with an encoder described above, all nodes in the original graph are transformed into low-dimensional continuous vectors ($\Phi \in \mathbb{R}^{|V|\times L}$) in the latent space, which are shown in the last column of \Cref{fig:vec_emb_workflow}. Each vector can be easily projected and visualized in 2D space as a point based on t-SNE~\cite{hinton2003stochastic} or PCA~\cite{ringner2008principal} techniques, where similar nodes in the original space will be close to each other in the latent space. Hence, similar nodes will be clustered together in the 2D space, and different types of nodes can be characterized by different clusters. The obtained node representations can provide effective task-independent node features for solving different downstream tasks (e.g., link prediction, node classification, node importance) with only simple traditional machine learning classifiers (e.g., SVM, random forest, k-means). Moreover, the low-dimensional vectors can be also further computed in different forms to feed into classifiers, such as concatenation, average or Hadamard product, etc. 

``Vector point-based graph embedding'' methods (summarized in \cref{tab:RW_compare} of \cref{sec:rw_emb_methods} in the Appendix) show great effectiveness for learning graph representation for diverse downstream tasks. However, the main drawback of such methods is that the most important network {\em uncertainty information} is not captured. Predicting the {\em uncertainty} for node embedding is particularly  important for quantitative analysis of large and complex network systems with diverse nodes and heterogeneous relations. 

\subsection{Gaussian distribution-based graph embedding}
\label{sec_3.2_gaussian_embedding}
Beyond the aforementioned deterministic vector point-based graph embedding methods in \cref{sec:section_3.1}, another type of emerging graph embedding methods (e.g., G2G~\cite{bojchevski2017deep}, KG2E \cite{he2015learning}, DVNE~\cite{zhu2018deep}) named ``Gaussian distribution-based graph embedding'' or ``probabilistic graph embedding'' holds great promise for stochastic graph embedding with important uncertainty estimation. Unlike the \textit{deterministic} ``vector point-based graph embedding'' method introduced in \cref{sec:section_3.1}, Gaussian distribution-based graph embedding enables learning of node embeddings as ``potential functions'' or ``continuous densities'' in latent space, which leads to two major advantages: a) it enables to effectively incorporate useful {\em unstructured attribute information} associated with each node, b) every node from the original graph is encoded as low-dimensional {\em multivariate Gaussian distributions} ($P_i \sim \mathcal{N}(\mu_i, \Sigma_i)$) in terms of the mean vector ($\mu_i \in \mathbb{R}^{L/2}$) and covariance matrix ($\Sigma_i \in \mathbb{R}^{L/2 \times L/2}$); the learned covariance term can provide extra free but important {\em uncertainty information}. The mathematical problem definition for Gaussian distribution-based graph embedding is shown in \cref{eq:mapping}. 
We summarize some recent works based on graph Gaussian embedding in \cref{tab:gauss_emb_surveytable} of \cref{sec:Gauss_emb_methods} in the Appendix. 

%%%%%%%%%%%%%%%%%%%%%%%%%%%%%%%%%%%%%%%%%%%%%%%%
The graph Gaussian embedding technique is inspired by the word2Gauss approach \cite{vilnis2014word}, which embeds words as Gaussian distributional potential functions in an infinite dimensional functional space. Therefore, each word is mapped into a ``soft region'' in latent space rather than a single point, which allows to explicitly model the uncertainty, entailment, inclusion, as well as providing a rich geometry for better quantification of the word type properties in the latent space. The aforementioned distinct properties using Gaussian embedding technique facilitate better representation of the word properties in latent space with uncertainty quantification. In addition, it is applicable for solving more complex graph representation learning tasks. In the following, we summarize the current graph Gaussian embedding methods from two graph types: 1) Gaussian embedding for knowledge graphs, 2) Gaussian embedding for attributed graphs. Generally, there are three main components for learning a graph Gaussian embedding model, including {\em triplet pair generation, energy function} and {\em loss function}.

\subsubsection{Gaussian embedding for knowledge graphs} 
\label{sec-kg_emb}
\begin{figure}[htbp]
    \centering
    \centering
    \includegraphics[width=.60\textwidth]{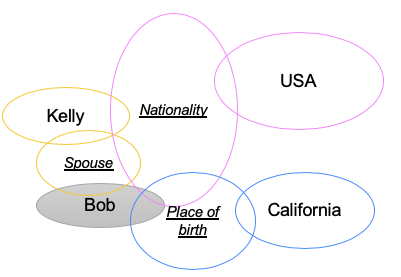}
    \caption{\textbf{Example of density-based knowledge graph Gaussian embedding in latent space.} Four entities (``Bob'', ``Kelly'', ``USA'', ``California'') and three types of relations (``Nationality'', ``Spouse'', ``Place of birth'') are shown in the diagram. Different colors indicate different facts/triplets, e.g., (Bob, spouse, Kelly) about the entity (``Bob''). Every fact (or triplet) consists of two nodes (head node and tail node) and a relation between these two nodes. Since Gaussian embedding is a density-based embedding approach, every entity (i.e., node) can be embedded into a soft region (i.e., ellipse) represented as a multivariate Gaussian distribution with mean and diagonal variances. The variances provide the corresponding uncertainty. Based on the density-based embeddings, we can infer that ``Bob was born in California''; on the other hand, the relation (``Spouse'') has lower uncertainty (i.e., smaller variances) than the relation (``Nationality'') while inferring a person (e.g., ``Bob'').} 
    \label{fig:uncertainty}
\end{figure}

He et al.~\cite{he2015learning} first proposed a Gaussian embedding model (KG2E) for learning stochastic knowledge graph (KG) representations in terms of Gaussian distributions, enabling  modeling of uncertainty of relations and entities in the knowledge graph using two benchmarks (i.e., WordNet~\cite{miller1998wordnet} and Freebase~\cite{bollacker2008freebase}). \Cref{fig:uncertainty} presents a density-based knowledge graph Gaussian embedding example for both ``entities'' (i.e., nodes) and ``relations'' (i.e., links) in the knowledge graph. Different entities and relations are transformed into a latent space of Gaussian distributions. The KG2E model is learned based on an energy-based learning framework~\cite{lecun2006tutorial}. In particular, it learns graph Gaussian representations (or embeddings) by using a \textit{margin-based triplet ranking loss} ($\mathcal{L}$) as shown in \cref{eq:max-margin-loss} with stochastic gradient descent (SGD) and the negative sampling technique. Hence, it can push the scores of positive triplets ($E_{pos}$) greater than the scores of negative triplets ($E_{neg}$) by a pre-defined margin ($\gamma$). 
%%%%%%%%%%%%%%%%%%%%%%%%%%%%%%%%%%%%%%%%%%%%%
%%%%%%%%%%%%%%%%%%%%%%%%%%%%%%%%%%%%%%%%
\begin{equation}
\mathcal{L} = \sum\limits_{(h, r, t)\in \tau}\sum\limits_{(h', r', t')\in \tau^{'}} max(0, E_{pos}(h, r, t) - \gamma + E_{neg}(h', r', t')). \\
\label{eq:max-margin-loss}
\end{equation}
%%%%%%%%%%%%%%%%%%%%%%%%%%%%%

In \cref{eq:max-margin-loss}, $(h,r,t) \in \tau$ denotes a {\em positive triplet} sample from the KG consisting of a relation ($r$) between the head entity ($h$) and tail entity ($t$), and $(h',r',t') \in \tau'$ denotes the corresponding {\em negative triplet} sample built by replacing $h$ or $t$ randomly (e.g., $(h', r, t)$ or $(h, r, t')$) using ``unif'' and ``bern'' techniques~\cite{wang2014knowledge}. In order to measure the similarity between the learned entity (node) and relation (link) embeddings in each positive and negative triplet sampled from the original KG, there are two different ways to define the energy function: a) Expected likelihood (``EL'') of inner product-based symmetric similarity, with the specific formula for denoting a positive triplet EL energy function $E_{pos}(h, r, t)$ shown in \cref{eq:EL-energy-func}; b) KL-divergence (asymmetric similarity), with the specific formula for denoting the positive triplet KL energy function $E_{pos}(h, r, t)$ shown in \cref{eq:kl-energy-func}. %
%%%%%%%% EL energy function%%%%%%%%%%%%%%
\begin{equation}
    \begin{split}
     E_{pos}(h, r, t) & = E_{pos}(P_e,P_r)\\
    & = \int_{x \in \mathbb{R}^{k_e}}\mathcal{N}(x; \mu_e, \Sigma_e) \mathcal{N}(x; \mu_r, \Sigma_r) dx\\
    & = \mathcal{N}(x; \mu_{e} - \mu_{r}, \Sigma_e + \Sigma_r)
    \end{split}
\label{eq:EL-energy-func}
\end{equation}
%%%%%%%% KL energy function %%%%%%%%%%%%%%
\begin{equation}
    \begin{split}
        &E_{pos}(h, r, t) = E_{pos}(P_e,P_r) = D_{KL}(P_e||P_r)\\
        &= \int_{x \in \mathbb{R}^{k_e}}\mathcal{N}(x; \mu_r, \Sigma_r)\log\frac{\mathcal{N}(x; \mu_r, \Sigma_r)}{\mathcal{N}(x; \mu_e, \Sigma_e)} dx\\
        &= \frac{1}{2}\{tr(\Sigma_r^{-1}\Sigma_e) + (\mu_r - \mu_e)^{T}\Sigma_r^{-1}(\mu_r - \mu_e)^{T} - \log\frac{det(\Sigma_e)}{det(\Sigma_r)} - k_e\},
    \end{split}
    \label{eq:kl-energy-func}
\end{equation}
%%%%%%%%%%%%%%%%%%%%%%%%%%%%%%%%%%%%%%%%
where $P_e \sim \mathcal{N}(\mu_h - \mu_t, \Sigma_h + \Sigma_t)$) denotes the transformation from head entity to tail entity of a triplet, with $\mu_{h} \in \mathbb{R}^{L/2}$ ($L$ is the embedding size) and $\mu_{t} \in \mathbb{R}^{L/2}$ denoting the mean vectors of head entity and tail entity embeddings; $\Sigma_h \in \mathbb{R}^{L/2\times L/2}$ and $\Sigma_t \in \mathbb{R}^{L/2\times L/2}$ denote the covariance matrices of the Gaussian embeddings for head entity and tail entity. The relation embeddings in latent space is denoted by $P_r \sim \mathcal{N}(\mu_r, \Sigma_r)$). In \cref{eq:kl-energy-func}, $tr(\cdot)$ indicates the trace and $\Sigma_r^{-1}$ refers to the inverse of the covariance matrix. Similarly, $E_{neg}(h', r', t')$ represents the negative triplet energy function that measures the similarity between entity and relation Gaussian embeddings (i.e., $P_e^{'}$ and $P_r^{'}$) for the corresponding negative triplets. It follows the same definition principles as shown in \cref{eq:EL-energy-func} and \cref{eq:kl-energy-func}.

\subsubsection{Gaussian embedding for attributed graphs}
Zhu et al.~\cite{zhu2018deep} proposed a deep variational autoencoder (AE)-based graph embedding model (DVNE) to learn Gaussian embeddings using one publication network benchmark (i.e., Cora~\cite{cora2000}) and three other social network benchmarks (i.e., Facebook~\cite{leskovec2012learning},  Flickr~\cite{mcauley2012image}). The DVNE model~\cite{zhu2018deep} is learned by optimizing a hybrid loss function combining a ranking loss (preserving 1st-order proximity) and a reconstruction loss (preserving 2nd-order proximity). DVNE employs the ``2nd-Wasserstein (W2) distance'' as shown in \cref{eq:W2-distance} 
%%%%%%%%%%%%%%%%%%%%%%
\begin{equation}
 E(P_i, P_j)^2 = W2(\mathcal{N}(\mu_i, \Sigma_i), \mathcal{N}(\mu_j, \Sigma_j))^2 = ||\mu_i - \mu_j||_2^{2} + ||\Sigma_i^{1/2} - \Sigma_j^{1/2}||_F^{2}
\label{eq:W2-distance}
\end{equation}
%%%%%%%%%%%%%%%%%%%%%%
to measure the similarity between the learned probabilistic Gaussian distributions for positive node pairs and negative node pairs, where $\mu$, $\Sigma$ stand for the predicted mean vector and diagonal covariance matrices corresponding to each positive or negative node pair ($v_i, v_j$). However, the DVNE model is only evaluated for undirected, non-attributed (plain) graph embedding problems.

\begin{figure}[htbp]
    \centering
    \includegraphics[width=.85\textwidth]{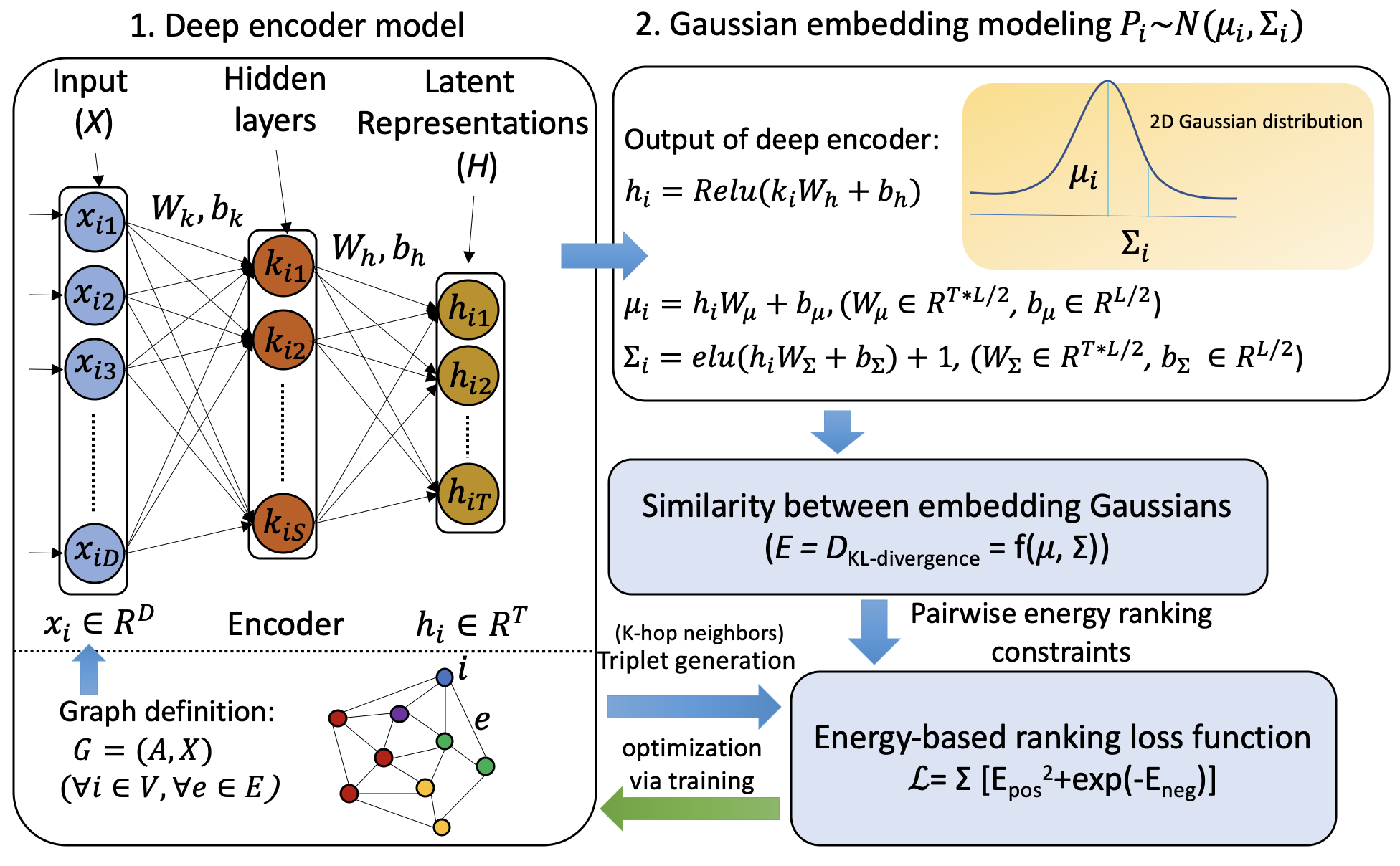}
    \caption{\textbf{Illustration of Gaussian distribution-based graph embedding framework.} First, node triplets are generated using the k-hop neighborhood method based on the adjacency matrix ($A$) of original graph $G$. Then, we input all node pairs and the corresponding attributes to deep encoder in 1) to learn node encodings ($p_\theta(h|x)$) with bottleneck of T. Finally, we use an ``uncertainty module'' to output the mean and variance vectors for Gaussian embeddings of each node. }
    \label{fig:gaussian_emb_workflow}
\end{figure}

Aiming to incorporate the important node attribute features for graph Gaussian embedding, Bojchevski et al.~\cite{bojchevski2017deep} proposed an inductive and unsupervised graph Gaussian embedding learning model, named Graph2Gauss (or G2G). It applies the ``multi-hop neighborhood sampling'' technique combined with a deep encoder (yet without decoder) to learn probabilistic graph embeddings (i.e., Gaussian distributions) in latent space for \textit{attributed} and \textit{directed/undirected} graphs. Specifically, let us consider a directed/undirected graph $G = (A, X)$ with $V$ and $E$ the corresponding vertex and edge sets, $A$ denotes the (symmetric or asymmetric) adjacency matrix of size $|V|\times |V|$, and $X$ is the attribute matrix of size $|V|\times D$. The main goal of \textit{Graph2Gauss} model is to project every node from the high-dimensional space into a latent space of multivariate Gaussian distributions with node attributes. For instance, the embedding of node $i$ ($P_i \sim \mathcal{N}(\mu_i, \Sigma_i)$) is a $L$-dimensional joint normal distribution with a mean vector ($\mu_i \in \mathbb{R}^{L/2}$) and a covariance matrix ($\Sigma_i \in \mathbb{R}^{L/2 \times L/2}$) in diagonal shape, where $L \ll D$. The main architecture of \textit{Graph2Gauss} contains four main elements: 1) Unsupervised node representation learning based on a deep encoder; 2) Node embedding modeling as Gaussian distributions; 3) Energy estimation for pairs of nodes in the embedding space; 4) Gaussian embedding learning by minimizing the energy-based loss, i.e., employing the square-exponential loss for the optimization of hyper-parameters of the deep encoder. The complete workflow is illustrated in \Cref{fig:gaussian_emb_workflow}, and we present details for the key components of the Gaussian embedding learning process below. 
%%%%%%%%%%%%%%%%%%%%%%
%%%%%%%%%%%%%%%%%%%%%%

\textbf{1) Node triplet generation:} Given a node $i$, its k-hop neighbors can be represented as $N_{ik}$  in \cref{eq:k-hop}, 
%%%%%%%%%%%%%%%%%%%%%%
\begin{equation}
    N_{ik} =\{j\in V|i \ne j, min(sp(i,j), K) = k\},
\label{eq:k-hop}
\end{equation}
%%%%%%%%%%%%%%%%%%%%%%
where $sp(i, j)$ denotes the shortest path between node $i$ and node $j$ (if $i$ and $j$ are not reachable it returns $\infty$); $K$ is the maximum considered distance, usually $K \ge 2$ enables capturing high-order proximity. A triplet sample usually consists of anchor, positive, and negative nodes [1]; a set of valid triplets can be represented as in \cref{eq:node_triplet}, 
%%%%%%%%%%%%%%%%%%%%%%
\begin{equation}
    D_t =\{(i, j_k, j_l)| sp(i,j_k) < sp(i,j_l)\}
\label{eq:node_triplet}
\end{equation}
%%%%%%%%%%%%%%%%%%%%%%
with $j_k \in N_{ik}$, $j_l \in N_{il}$ and $k < l$. Thus, for the triplet $(i, j_k, j_l)$, node $i$ is more similar to node $j_k$ than node $j_l$, and the node pair $(i, j_k)$ denotes one positive pair, while the node pair $(i, j_l)$ denotes one negative pair, which are generated for the energy-based ranking loss construction. Moreover, a ``node-anchored sampling'' strategy~\cite{bojchevski2017deep} provides an effective way for triplet generation that can help reduce the computational complexity in a large graph.

\textbf{2) Network structure preservation via personalized ranking:} In order to capture the network structure properties at a multi-scale level for graph embedding, personalized ranking of energy (similarity) constraints in \cref{eq:net_structure} 
%%%%%%%%%%%%%%%%%%%%%%%%%%%%%
\begin{equation}
\begin{array}{ll}
  E(P_i, P_{k_1}) < E(P_i, P_{k_2})  < ... < E(P_i, P_{k_K}),\\
  \forall{k_1}\in N_{i1}, \forall{k_2}\in N_{i2}, ,...,\forall{k_1}\in N_{iK}
\end{array}
\label{eq:net_structure}
\end{equation}
%%%%%%%%%%%%%%%%%%%%%%%%%%%%%
are imposed to the latent node embeddings. That is, the respective energy (or distance) between embeddings of node $i$ and each node in its $k$-hop neighborhood (``positive energy'') should be lower than the one between embeddings of node $i$ and its $(k+1)$-hop neighbors (``negative energy''). Here, $E(P_i, P_j)$ denotes the energy function between learned Gaussian distributions $(P_i, P_j)$ for nodes $i$ and $j$.

\textbf{3) Similarity measure for embedding Gaussians:} \textit{Graph2Gauss} employs the asymmetric KL-based energy function (same as \textit{KG2E} in \cref{eq:kl-energy-func}) to measure the distance between the learned nodes' Gaussian distributions $(P_i, P_j)$ and score the positive and negative triplets. Here $(i, j)$ can be either positive node pairs or negative node pairs. Currently, the most commonly used similarity metrics include the symmetric expected likelihood (EL), the Jensen-Shannon divergence (JS), the asymmetric KL-divergence (\textit{KL}), and the $p-th$ Wasserstein distance $(Wp)$. Different from the first two metrics, $KL$ and $Wp$ can be also used to handle {\em directed graph} embeddings while at the same time preserving the transitivity of nodes. Obtaining smaller energy ($E$) represents that the nodes’ embedding Gaussians are more similar or closer to each other.

\textbf{4) Energy-based ranking loss function:} Based on the aforementioned constraints on the respective energy between latent embeddings of adjacent k-hop neighbors of each node (e.g., 2-hop vs. 3-hop of each node), in order to learn a graph embedding that satisfies the constraints, the energy-based learning approach~\cite{lecun2006tutorial} is used. The \textit{Graph2Gauss} model is trained with an energy-based ranking loss $(\mathcal{L})$ over the sampled triplets $(D_t)$ for penalizing the ranking errors, such that positive energy $E_{ij_k}$ terms are always lower than negative energy $(E_{ij_l})$. The ranking-based loss ($\mathcal{L}$) usually consists of two parts: positive node pair energy ($E_{ij_k}$) and negative node pair energy ($E_{ij_l}$). ``Margin-based ranking loss'' is a frequently used loss function for graph embedding learning, however, the margin has to be manually selected before training. Thus, the ``square-exponential loss'' representing negative pair energy as an exponential term has better performance in penalizing the ranking error automatically; the specific formula is given in \cref{eq:square-exp-loss}.
\begin{equation}
   \mathcal{L} = \sum\limits_{(i,j_k,j_l) \in D_t}(E_{ij_k}^2 + exp^{-E_{ij_l}}), k<l. 
\label{eq:square-exp-loss}
\end{equation}
The uncertainty-aware Graph2Gauss model can learn a high-dimensional graph in a probabilistic latent space incorporating both graph structure and auxiliary attributes information. Moreover, multi-scale graph structure properties can be preserved during graph embedding by performing k-hop neighborhood sampling and energy-based personalized ranking. Additionally, the learned important uncertainty term can be used for multiple aspects: 1) Quantification of the neighborhood diversity for every node (i.e., the number of distinct classes in a node's neighborhood); and 2) Discovering the intrinsic latent dimensionality of the graph (i.e., $\approx L$); see an example in \cref{sec:Applications}. In particular, when the dimensions with high average uncertainty are removed, the link prediction performance remains stable and without a decrease. More importantly, the G2G model is robust to the number of training edges, the number of labeled nodes, as well as the hyper-parameters of the neural networks.

\subsection{Dynamic Graph Embedding}
\label{sec_3.3}
In real applications, many networks exhibit dynamics and evolve over time in terms of growing or shrinking nodes (or links), altering relations or edge weights. Dynamic graphs present a lot of benefits for representing the ubiquitous interactive networks, e.g., evolving gene-protein networks, financial transaction graphs, traffic-flow networks,
%[PreME dataset], 
longitudinal citation networks, and dynamic social communication networks. Recently, studies on dynamic graph embedding have attracted considerable attention due to its capability of capturing the underlying evolving patterns (or embeddings) for each node at different time steps, hence facilitating more accurate forecast of the future connections between nodes~\cite{goyal2020dyngraph2vec}. 

\subsubsection{Definitions of dynamic graphs}
\begin{figure}[htbp]
    \centering
    \includegraphics[width=.70\textwidth]{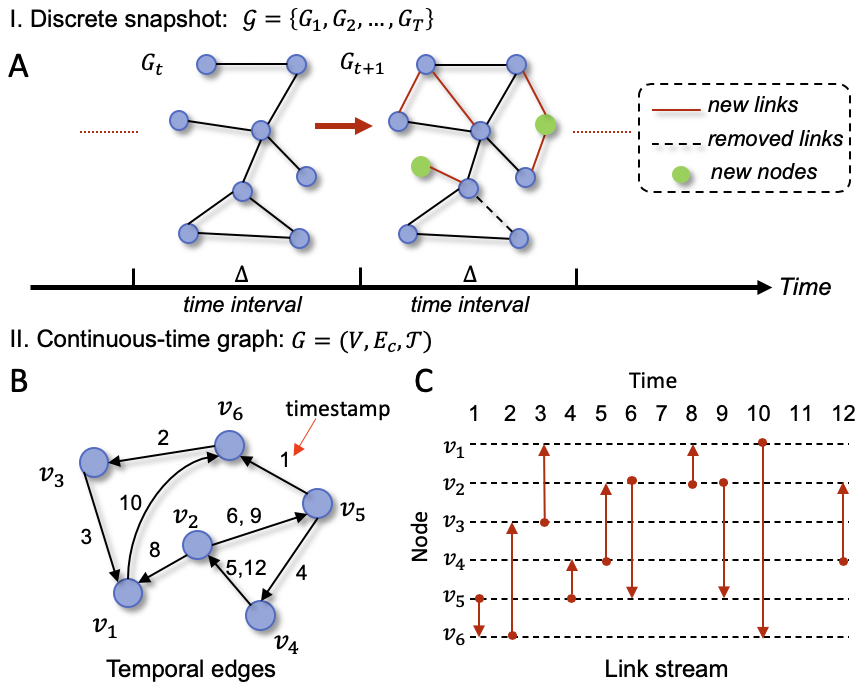}
    \caption{\textbf{Different representations of a dynamic graph.} (A) Modeling a discrete-time dynamic graph $\mathcal{G}$ as a series of discrete snapshots at different time intervals ($\triangle$), i.e., $\mathcal{G} = \{G_1, G_2, ..., G_T\}$), which contains $T$ static snapshots; $G_{t+1}$ is the next snapshot of snapshot $G_t$ with five new added links (red solid lines), two new added nodes (green dots) and a removed edge (black dashed line). (B) Modeling a continuous-time dynamic graph (CTDN) $\mathcal{G}$ through assigning each edge with multiple timestamps based on a edge-time mapping function ($\mathcal{T}$), i.e., representing a continuous time dynamic graph as $G=(V, E_c, \mathcal{T})$, where $V$ is the node set, $E_c$ indicates the temporal edge set with each element as a triplet, e.g., $(v_2, v_5, 6), (v_2, v_5, 9)$ represent two temporal edges starting from node $v_2$ to $v_5$ at time $t = 6, 9 \in \mathcal{T}(v_2, v_5)$. (C) Link stream is another way to represent a continuous time dynamic graph; the x-axis indicates the time instants while the y-axis represents all the nodes in dynamic graph.}
    \label{fig:dyn_graph}
\end{figure}

Generally, a dynamic graph ($\mathcal{G}$) can be mathematically defined in two different ways: 
\textbf{1) Discrete snapshots:} A dynamic graph can be approximated as a sequence of discrete static graph snapshots with equal time intervals ($\mathcal{G} = \{G_1, G_2,..., G_T\}$); see the snapshot-based example in \Cref{fig:dyn_graph}(A), where $G_t = (V_t, E_t), t \in [1,T]$ denotes the $t-th$ snapshot with the node set $V_t$ and the edge set $E_t$, while $T$ refers to the number of snapshots. Notably, the problem of using snapshot definition is that it assumes coarse-grained global evolution changes. Hence, the important continuous time-varying graph evolution properties cannot be captured. 
%%%%
\textbf{2) Continuous-time graphs:} There are two ways to model a graph as a continuous-time changing graph. 
\textit{i) Temporal edge-based}: a continuous-time changing graph $G=(V,E_c,\mathcal{T})$ can be denoted by a node set $V$, a temporal edge set $E_c \subseteq V \times V \times \mathbb{R}^{+}$ and a time mapping function $\mathcal{T}: E\rightarrow \mathbb{R}^{+}$ ($\mathbb{R}^{+}$ is the time domain) for every edge, and every edge is labeled by time (see an example in \Cref{fig:dyn_graph}(B)). Hence, the important temporal sequence order and flow information in the temporal graphs can be captured in a fine-grained manner. Notably, sampling valid walks from a dynamic graph has to follow the increasing time order, e.g.,  ``$v_1\rightarrow v_6\rightarrow v_3$'' is not a valid walk in the \Cref{fig:dyn_graph}(B) since edge ($v_6,v_3$) is in the past of edge ($v_1,v_6$). 
\textit{ii) Link stream-based}: another alternative continuous time graph representation is ``link stream'', and \Cref{fig:dyn_graph}(C) shows the corresponding link stream representation for the dynamic graph in \Cref{fig:dyn_graph}(B). The vertical axis represents all the nodes of the dynamic graph and the horizontal axis shows different time instants, while the red arrows correspond to the link streams at different time points. As such, the continuous time dynamics can be fully represented in a single graph. Additionally, continuous time dynamic graph representations (temporal edges and link streams) facilitate sampling temporally valid walks from dynamic graphs, hence enabling learning of more meaningful and accurate graph representations.
%%%%%%%%%%%%%%%%%%%%%%%%
%%%%%%%%%%%%%%%%%%%%%%%%
\subsubsection{Problem settings for dynamic graph embedding}
Based on the snapshot definition above, dynamic graph embedding can be defined as a time-series of mappings ($\mathcal{F} = \{f_1, f_2, ..., f_T\}$) corresponding to different snapshots~\cite{goyal2018dyngem}, where $f_T$ is the embedding of the snapshot ($G_T$). Moreover, the graph structure properties and temporal dynamics are preserved in the latent space. For continuous time dynamic networks (CTDNs), the embedding problem can be formulated as learning of a mapping function $f: V\rightarrow \mathbb{R}^L s.t. L \ll V$, such that every node is embedded into low dimensional space as a $L$-dimensional time-dependent vector (see related works in ~\cite{nguyen2018dynamic}).

There are three key challenges in dynamic graph embedding: i) \textit{How to learn a stable graph embedding ($\mathcal{F}$)}, such that the mappings are also similar when the adjacent snapshots only exhibit subtle dynamics/changes. Specifically, the stability of $\mathcal{F}$ can be evaluated by a ``stability constant'' measure ($\mathcal{K}_s(\mathcal{F})$ in \cref{eq:stability}) 
%%%%%
\begin{equation}
\begin{array}{ll}
\mathcal{K}_s(\mathcal{F}) = \underset{\tau, \tau^{'}}{\max}|S_r(F;\tau)-S_r(F;\tau^{'})|\\
\text{where} \;\;\; S_r(F; t) =\frac{||f_{t+1}(V_{t})-f_{t}(V_{t})||_F}{||f_{t}(V_{t})||_F}/\frac{||S_{t+1}(V_{t})-S_{t}(V_{t})||_F}{||S_{t}(V_{t})||_F},
\end{array}
\label{eq:stability}
\end{equation}
%%%%%%
by computing the maximum ``relative stability'' difference between different neighboring snapshot pairs ~\cite{goyal2018dyngem}. The \textit{relative stability} $(S_r(\mathcal{F};t))$ is denoted by the ratio of the relative difference between embeddings to that of the relative difference between adjacency matrices of two neighboring snapshots. In particular, $f_{t}(V_t) \in \mathbb{R}^{|V_t|*d}$ denotes the node embedding matrix for all nodes ($V_t$) in the $t-th$ snapshot, where $d$ refers to the embedding size. Similarly, $S_{t}(V_{t}), S_{t+1}(V_{t})$ denote the adjacency matrices for two neighboring snapshots (see the corresponding definitions in \cref{eq:stability}). A small stability constant indicates that the embedding is more stable.

ii) \textit{How to effectively and efficiently update the node embedding} with topological graph evolution over time. iii) \textit{How to scale it} to large-scale dynamic network embedding.

To tackle these problems, some dynamic graph embedding models have been developed and are listed in \cref{tab:dynGEM} in the Appendix, which contains five different categories: 1) Matrix factorization-based, 2) SkipGram-based, 3) Autoencoder (AE)-based, 4) Graph convolutional networks (GCN)-based, and 5) Generative adversarial network (GAN)-based. We refer the interested readers to \cref{sec:dyn_GEM} in the Appendix for references to these methods.
%%%%%%%%%%%%%%%%%%%%

\section{Applications}
\label{sec:Applications}
An increasing number of diverse applications have employed the aforementioned deep learning-based graph embedding methods across different fields due to their inductive and unsupervised non-linear graph mapping function learning property. High-dimensional and large-scale graphs can be encoded as continuous, low-dimensional graph embeddings in the latent space, while the latent space graph patterns (i.e., embeddings) can be readily used for solving diverse downstream graph analytic problems. For example, detecting an anomaly in social networks~\cite{tan2019deep,ding2019deep} or financial networks \cite{bruss2019deeptrax}, analyzing the non-pharmacological cognitive training effects using functional brain network embedding patterns~\cite{xu2019gaussian} and to predict early stages of Alzheimer's disease \cite{xu2020graph}, mining biochemical multi-scale structures in biochemical graphs \cite{wang2020gaussian}, finding functional modules or sub-compartments in genomic networks~\cite{ashoor2020graph}, etc. In the following, we highlight some representative applications based on four graph types: 1) social networks, 2) citation networks, 3) brain networks, 4) genomic networks. Moreover, all the public benchmark datasets introduced in the following can be obtained using the easy-to-use APIs provided in public libraries (e.g., PyTorch Geometric library~\cite{Fey/Lenssen/2019}, Deep Graph Library (DGL) package~\cite{wang2019deep}).

% 1. social network application
\subsection{Social network applications}
\begin{figure}[htbp]
    \centering
    \includegraphics[width=.85\textwidth]{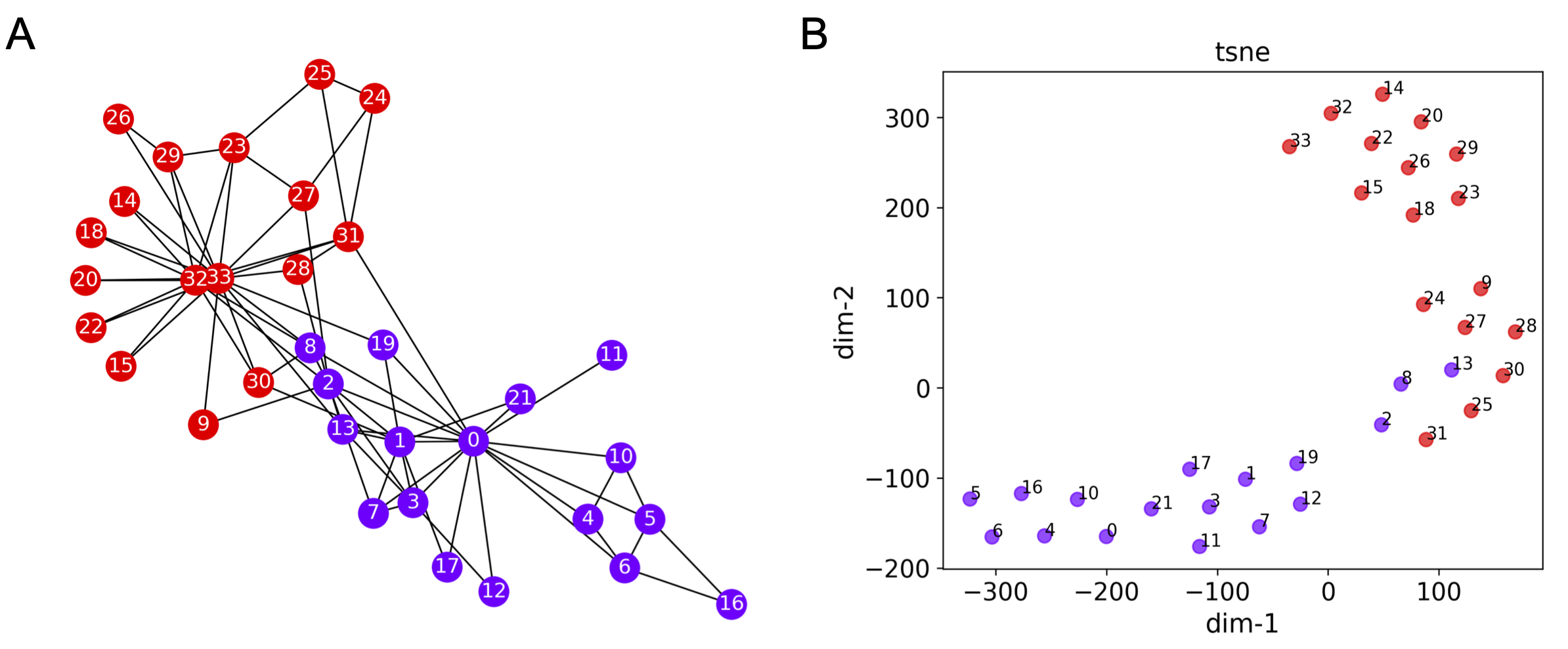}
    \caption{\textbf{Example of applying graph embedding method for community detection in the karate club dataset~\cite{zachary1977information}}. (A) Visualization of the Zachary's karate club network~\cite{zachary1977information}; it contains 34 nodes and 154 undirected and unweighted edges with nodes of the same color belonging to the same ground-truth community. (B) Scatter plot for the low-dimensional node embedding vectors obtained by the DeepWalk method~\cite{perozzi2014deepwalk}; each node embedding vector with a dimension of 128 was projected into 2D euclidean space via the t-SNE method~\cite{hinton2003stochastic}. For the interested readers, the detailed implementation is described in \cref{ipl_kc} in the Appendix.}
    \label{fig:social_net}
\end{figure}

With the increasing user-generated content collected from public social media sites and mobile apps, there are a lot of challenging social network mining tasks (e.g., forming conclusions about users, acting upon the information, advertising to users, etc.). Graph embedding techniques can provide a very useful and effective graph representation learning tool for solving these challenging problems. For example, the Zachary's karate club network~\cite{zachary1977information}) consists of social friendships (links) between 34 members (nodes) of a university karate club. In order to predict the membership for each member, Perozzi et al. adopted a deterministic graph node embedding method (called ``DeepWalk''~\cite{perozzi2014deepwalk}) to learn a low-dimensional vector representation for each node, and then the relationships between different nodes can be captured by the distance (i.e., the metrics in \cref{tab:similarity_embedding_space}) between node embedding vectors in the latent space, see example in \Cref{fig:social_net}. This is a relatively simple benchmark for the interested reader to practice so we present implementation details in \cref{sec:implement}.

%large graph embeddings
In addition, studies also demonstrated the emerging graph embedding methods performed very robust and in particular computationally efficient for analyzing large-scale attributed social networks (e.g., Facebook, Github, Friendster and Twitter network datasets~\cite{snapnets}). For example, the authors in \cite{zhang2017user} leveraged a transfer learning technique in the traditional SkipGram-based graph embedding model for learning node embeddings of attributed large-scale networks; the embeddings were subsequently used as latent features for different machine learning tasks, e.g., multi-class classification, regression, etc. 
%%%%%%%

% 2. citation networks
\subsection{Citation network applications}
Given the vast amount of publications on multidisciplinary fields, a citation network can be used as an effective way to represent the complex citation relationships (``directed links'') among the publications (``nodes'') from different scientific research domains. In addition, citation network nodes are usually assigned with attributes (e.g., publication date, paper version, etc.), which provide useful auxiliary node features for better characterizing the heterogeneous relationships among network nodes. Advanced graph embedding methods enable us to solve diverse citation network analytics problems more quantitatively and efficiently. 

\begin{figure}[htbp]
    \centering
    \includegraphics[width=1.0\textwidth]{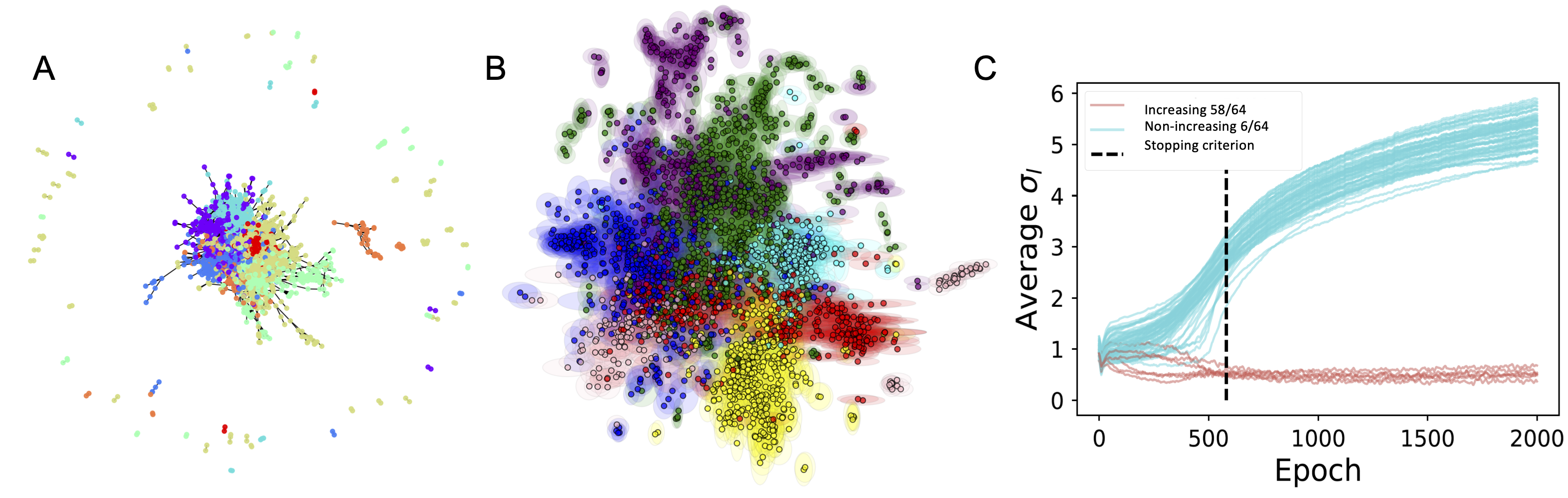}
    \caption{\textbf{Example of applying stochastic graph embedding method for node classification in CORA-ML dataset~\cite{bojchevski2017deep}}. (A) Visualization of the CORA-ML network~\cite{bojchevski2017deep}; it consists of 2995 nodes, each node has 2879 attributes, and all nodes are divided into 7 classes, which are plotted using different colors. (B) Scatter plot for the stochastic graph embedding results generated by the graph2gauss method~\cite{bojchevski2017deep}. The oval shapes around the points denote the uncertainty. (C) Average uncertainty per dimension versus the number of epochs during training. The embedding size is $L = 2 + 2$; implementation details can be found in \cref{ipl_cml} in the Appendix. }
    \label{fig:uncertainty_benefits}
\end{figure}

For example, in paper~\cite{bojchevski2017deep}, the authors adopted a generative network-based model to encode every node of the directed and attributed citation networks into a multivariate Gaussian distribution with mean and variance vectors, which can effectively capture uncertainty information. Moreover, the advantages of quantifying uncertainty using the variance output has been demonstrated in \cite{bojchevski2017deep} by analyzing two subset datasets (CORA and CORA-ML) extracted from the original ``Cora'' citation network dataset~\cite{cora2000}. 
% CORA citation network consists of  scientific publications classified into one of seven classes. The citation network consists of 5429 links. Each publication consists of 1433 unique words.
In addition to the physical interpretation associated with possible conflicts in link prediction or node classification, uncertainty quantification (UQ) can also be used to estimate neighborhood diversity and most importantly to infer the intrinsic latent dimensionality of a graph. In particular, more homogeneous nodes have small variance compared to more heterogeneous nodes whose adjacent neighbors may be a member of multiple classes, leading to a more uncertain embedding. On the issue of dimensionality, we show in \Cref{fig:uncertainty_benefits} an important result from \cite{bojchevski2017deep} for the Coral-ML dataset.  The plot describes two sets of lines (red and blue, 64 in total) with each line representing the average variance for a give dimension $l$ over all nodes as a function of the epoch number for a link prediction validation test. It is interesting to see that the lines split clearly into two distinct sets fairly quickly as training continues: a small subset of lines with low uncertainty that asymptotes a small variance constant value, and another one with increasing variance, which obviously denotes the unstable dimensions. The authors of \cite{bojchevski2017deep} also observed that the effective dimensionality ($L$ = 6) observed using this approach is very close to the number of ground-truth communities (7); this is also something we observed in our own work in the two neuroscience tasks that we discuss below.
%%%%%%%%%%%%%%%%%%%%%%%%%%
%%%%%%%%%%%%%%%%%%%%%%%%%%

% 3. neuroscience brain networks
\subsection{Brain network applications}
The brain system consists of spatially distributed but functionally linked specialized brain elements (neurons, neural assemblies, or brain regions), i.e., ``nodes'',  which continuously share information with each other and form a complex ``brain network'' (or connectome). Graph theory-based brain network analysis has received extensive attention in the past few decades, however, few studies have focused on latent node representation learning for brain networks. Moreover, less attention has been paid to conducting node-wise quantitative comparisons among different brain networks~\cite{mheich2020brain}. Graph embedding methods can provide a powerful network representation learning tool and can effectively tackle the aforementioned key issues with hierarchical and quantitative brain network analysis. In particular, graph embedding learning is applicable to multi-modality brain network analysis and bears great potential in characterizing both latent link-level and node-level features in the brain networks, which can be further used for identifying vital multi-scale network markers for a wide variety of brain disorders. 

For example, Rosenthal et al.~\cite{rosenthal2018mapping} proposed a predictive model that employed the node2vec~\cite{grover2016node2vec} method to learn vector-based embeddings for cortical regions in the brain networks, which were then used for link inference between brain network structure and function. Their experimental results demonstrated an important advantage of the latent brain network embeddings, i.e., that it can simulate the effects of localized network lesions on the global pattern of functional connectivity. 

\begin{figure}[htbp]
    \centering
    \includegraphics[width=1.0\textwidth]{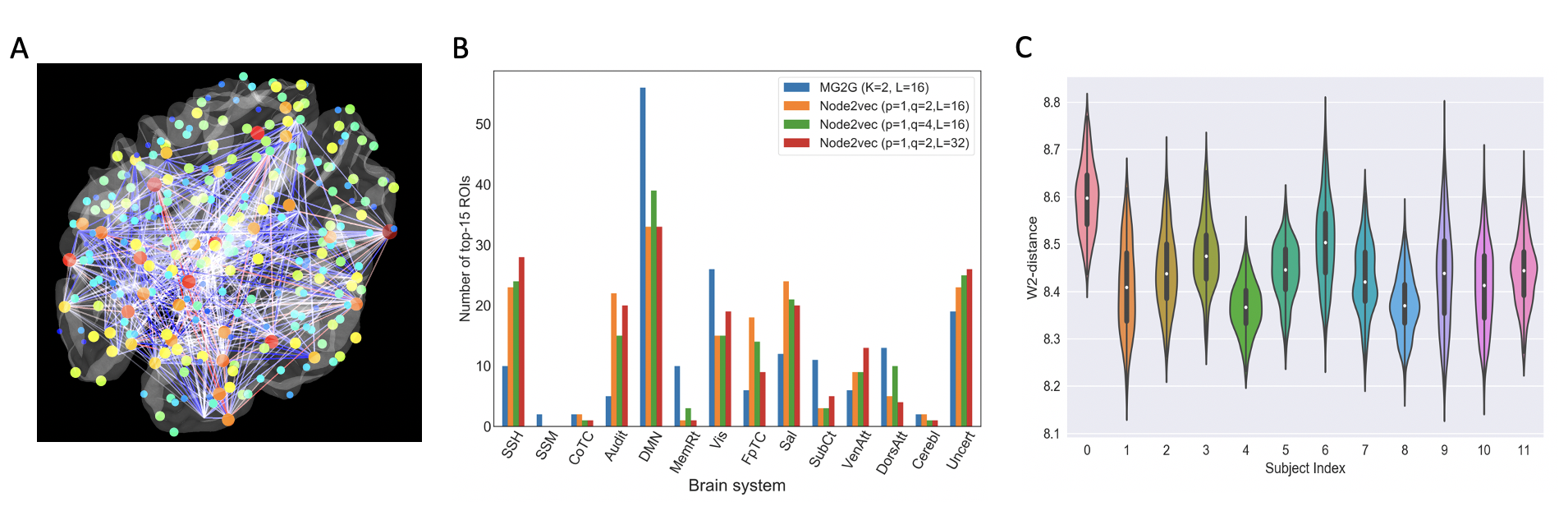}
    \caption{\textbf{Example of applying stochastic (MG2G) and deterministic (node2vec) graph embedding methods for cognitive training effects using fMRI brain networks}. (A) Visualization of the top 50 affected brain regions with the corresponding brain connectivity due to multi-domain cognitive training (MDCT)  detected by fMRI brain network embedding for one of the aMCI patients. (B) and (C) are from~\cite{xu2020new}. (B) illustrates the quantification of functional and system-level changes for 12 patients before and after MDCT intervention based on MG2G (blue) and node2vec~\cite{grover2016node2vec} (yellow, green, and red correspond to different node2vec parameters); (C) MG2G result: Violin plot for the W2-distance distributions and probability densities of all 264 regions w.r.t. different patients (the embedding size L = 16).}
    \label{fig:brain_net}
\end{figure}

In another application, in order to analyze the subtle multi-domain cognitive intervention effects for amnestic mild cognitive impairment (aMCI) patients using fMRI data, the authors of \cite{xu2020new} presented a functional brain network analysis method based on the multi-graph unsupervised Gaussian embedding method (MG2G), which applied graph weights to sample node neighbors. The model can encode network nodes as multivariate Gaussian distributions (``embeddings''), and employs the Wasserstein distance for quantitative evaluation of brain network changes (at region-level, system-level and subject-level) after 3-month of multi-domain cognitive training (MDCT), i.e., a non-pharmacological intervention. The embedding variances provide extra uncertainty quantification for the latent node embeddings, which lead to better patient-specific evaluation for the cognitive training effects as well as the extra benefit of identifying the intrinsic dimensionality of the brain network, which is approximately equal to the number of communities (14) in the brain. These advantages are apparently superior to the deterministic embeddings applied in the previous study~\cite{rosenthal2018mapping}. \Cref{fig:brain_net} presents quantitative results for evaluating the MDCT effects for aMCI patients. In \Cref{fig:brain_net} (A), different colors of the ``dots'' (i.e., 264 brain regions in the fMRI brain network) correspond to the computed Wasserstein-2 (W2) distance values, which measures the learned low-dimensional stochastic resting state fMRI brain network embedding distance before and after cognitive intervention in the latent probabilistic space. ``Links'' represent the sampled brain connectivity among the detected the top 50 affected brain regions (mostly concentrated in the \em{default mode}, \em{visual}, and {\em memory retrieval} functional brain systems). In \Cref{fig:brain_net} (B) we compare a stochastic (MG2G) and a deterministic (node2vec) graph embedding method; the plot shows the top 15 most affected brain regions for all patients measured by the W2-distance, and identified with the brain systems they belong to. \Cref{fig:brain_net} (C) shows a violin plot of the W2-distance distributions and probability densities of all 264 regions for all 12 different patients included in this aMCI study~\cite{xu2020new}.

Beyond the aforementioned cognitive training effect assessment application, using graph embedding for brain network representation learning can also provide a promising tool for characterization of Alzheimer's disease (AD) progression~\cite{xu2020graph}. Specifically, the obtained brain network embeddings can be used as latent features to feed into classic classifiers (e.g., random forest, SVM, SVD, etc.) for identifying different early stages in AD progression using Magnetoencephalography (MEG) brain networks. The learned high-informative latent brain network embeddings can also be effectively used for quantitatively identifying specific brain regions with network alterations related to mild cognitive impairment (MCI). 
%%%%%%%%%%%%%%%%%%%%%%%%%%
%%%%%%%%%%%%%%%%%%%%%%%%%%
 
% 4. Gene networks
\subsection{Genomic network applications}
The dimension reduction property of graph embedding techniques, in addition to computational expediency, it can also tackle the noise and incomplete problems existing in large molecular network datasets. Here, we present two different applications, where two types of graph embedding techniques (deterministic and stochastic) are used for genomic network analysis.

%%%%%%%%%%%%%%%%%%%%%%%%%%
\begin{figure}[htbp]
    \centering
    \includegraphics[width=1.0\textwidth]{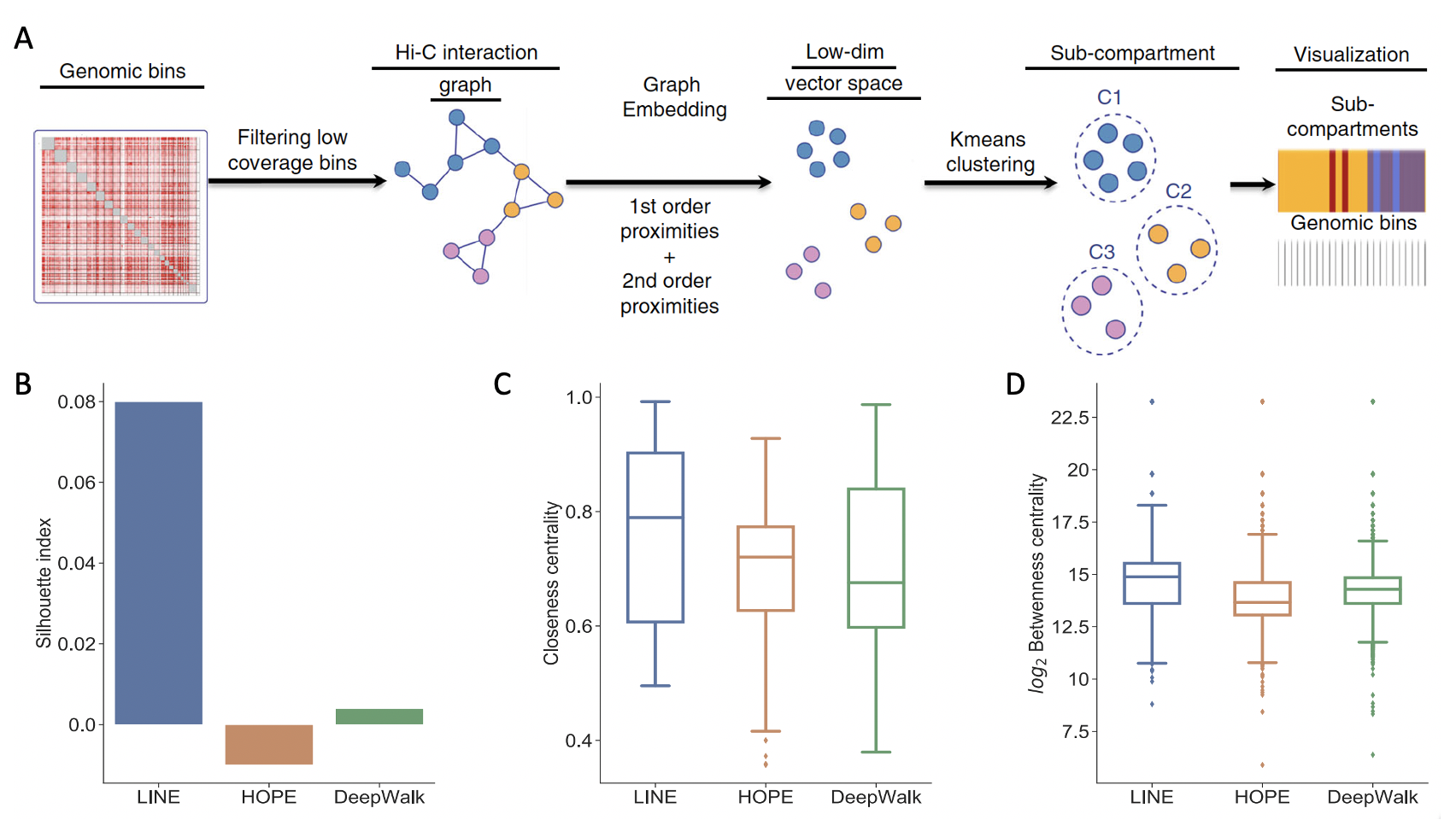}
    \caption{\textbf{Example of applying the deterministic LINE graph embedding method for sub-compartment identification using Hi-C chromatin interaction data}. (A) Workflow from left to right proposed by~\cite{ashoor2020graph}, showing a normalized Hi-C interchromosome matrix used to construct the interaction graph. The nodes with the same color denote genomic bins from the same chromosome. LINE projects the graph into a lower-dimensional space for a subsequent k-means clustering step.  Comparative performance of LINE against two other graph embedding methods HOPE and DeepWalk. The plots show the Silhouette index (B), the closeness centrality (C), and the betweenness centrality (D). The line in each box represents the median of the samples while the other lines show their 75th and 25th percentiles. Both figures adopted from~\cite{ashoor2020graph}.}
    \label{fig:gene_net}
\end{figure}
%%%%%%%%%%%%%%%%%%%%%%%%%%

 One of the fundamental open problems in computational biology is how the genome is organized in the nucleus of a cell in three-dimensional hierarchical sub-compartments as these sub-regions exhibit specific epigenomic signatures. Chromatin plays a key role in defining this spatial organization, and hence studying chromatin interactions (Hi-C) can shed light into this fundamental question. The authors of \cite{ashoor2020graph} employ the LINE graph embedding method~\cite{tang2015line} and unsupervised learning to predict the sub-compartments in the nucleus based on data describing Hi-C chromatin interactions obtained by a profiling technique combining massively parallel sequencing and proximity-based ligation. The workflow consists of multiple steps and begins by preprocessing the Hi-C interaction data to output a normalized intercromosome matrix based on which the interaction graph is constructed. A schematic of this workflow is shown in \Cref{fig:gene_net} (A). Areas (genomic bins) with the same chromosome are represented by the same color nodes. Based on the constructed interaction graph, the graph embedding algorithm projects the graph into a lower-dimensional space for k-means clustering to predict the various hierarchical sub-compartments. The method proposed in \cite{ashoor2020graph} based on LINE embedding outperforms the hidden Markov model (HMM) for sub-compartment prediction, and in fact it predicts a larger number (9 versus 5) of sub-compartments. It also outperforms two other graph embedding methods that we described in this review paper, namely HOPE~\cite{ou2016asymmetric} and DeepWalk~\cite{perozzi2014deepwalk}, as shown in \Cref{fig:gene_net} (B-D). Genomic regions that map to the same sub-compartment based on the graph embedding method are spatially close to each other. In the study of \cite{ashoor2020graph} this is measured quantitatively using the network topology features, including the network centrality (closeness and betweenness) as well as the Silhouette indices, which reflect the consistency within each data cluster. The Silhouette index ranges from $-1$ to 1, with a higher value indicating better clustering performance. The LINE method \cite{tang2015line} improves the Silhouette index compared to HMM and is also superior to HOPE and DeepWalk with respect to all three metrics as shown in \Cref{fig:gene_net} (B-D).

In the second application, we describe a novel use of Gaussian embedding for gene set member identification.
Gene sets give rise to big data given the recent advances in high-throughput biological data, however processing such data to gain biological insights has been hindered by the lack of appropriate methods to make them useful for downstream tasks.
The authors of~\cite{wang2020gaussian} employed Gaussian embedding to formulate a new method (called ``set2Gauss'') to encode each gene set as a multivariate Gaussian distribution in the latent space. 
Specifically, set2Gaussian requires as an input a biological network (e.g., a protein-protein network) as well as 
a group of gene sets. Each gene is represented as a single point while each gene set is modeled via a multivariate
Gaussian distribution. The projection from the graph to the low-dimensional vector space is based on the 
proximity of the genes in the protein-protein interaction network. Genes in the same set may have different functions,
but such differences have been ignored in the standard average embedding methods. 
This functional diversity in each gene set is naturally modeled in set2Gauss by the uncertainty of each dimension
in the projected vector. In~\cite{wang2020gaussian}, Wang et al. demonstrate for three different large-scale gene set groups that set2Gauss can capture differences between gene sets that standard approaches (e.g., mean pooling) would ignore.
Overall, the detailed analysis of~\cite{wang2020gaussian} demonstrates that by using the graph Gaussian embedding approach for projecting the gene sets into latent space as multivariate Gaussian distributions, it can effectively model the {\em uncertainty} and capture the functional diversity within a gene set, i.e., inherently capture the properties of redundancy, size and coverage of gene sets.

\section{Summary and Discussion}
\label{sec:Summary}
%
% Graph analytics - why use it and for what; summarize preliminaries of Section 2.

Graph analytics has become an indispensable tool in research as well as in the industrial sector in recent years as it can deal effectively with large and noisy datasets across different application domains. Industrial systems may employ graphs with over 50 million nodes and over a billion edges, hence it is very important to employ and further develop low computational complexity algorithms. Graph embedding is an effective approach for transforming graphs into low-dimensional vector representations for individual vertices in networks. In this Review, we have provided introductory material for the mathematical formulation of the graph embedding problem, including preliminaries (i.e., basic graph notations and definitions) as well as specific graph embedding settings for deterministic and stochastic graph embedding problems. 

% Past versus current graph embedding methods
An important high-level task in graph analytics is to transform a high-dimension original graph to a more compatible representation for efficient downstream processing tasks, e.g., node classification, link prediction, community detection, etc. A key milestone towards this goal in the modern era was the seminal paper on Isomap~\cite{tenenbaum2000global}, a nonlinear dimensionality reduction method, which introduced the geodesic distance instead of the Euclidean distance used in the classical Principal Component Analysis (PCA). Isomap can be applied to a wide range of datasets but in the last twenty years since the first publication of Isomap, there have been great developments that have reduced the computational complexity and have increased the accuracy in diverse network applications as we demonstrated in the previous section \ref{sec:Applications}. For example, the computational complexity of Isomap is $\BigO{L^2 M}$ for a graph with $N$ nodes and $M$ ($M\ge N$) edges with embedding size $L$. In contrast, the stochastic G2G method we highlighted in the previous section has linear computational complexity, i.e., $\BigO{N}$. Hence, one can work with very large graphs with $N=20,000$ nodes as in the CORA and PUBMBED datasets at reasonable computational cost on a single GPU. Moreover, in modern graph embedding methods, we can quantify uncertainty, which can have useful interpretations as in the aforementioned example in genetic network analysis in \cref{sec:Applications}.

Broadly, we can classify the graph embedding methods in three major categories: matrix factorization-based methods, random walk-based methods, and neural network-based methods. Matrix factorization-based methods, e.g. HOPE, cannot easily scale up to large network embeddings; their computational complexity is similar to that of Isomap. Hence, the focus of this Review has been on random walk-based methods and neural network-based methods, e.g., LINE, DeepWalk, node2vec, G2G, etc., which are characterized by computational complexity of $\BigO{L M}$,  $\BigO{L N}$ or $\BigO{N}$. Moreover, in all these methods only unsupervised learning is required. In the Appendix \cref{sec:rw_emb_methods}, we summarize the main characteristics of some of these methods along with their specific computational complexity. We also  provide implementation details for both deterministic and stochastic graph embedding methods using the relatively simple datasets of karate club as well as the CORA-ML (see \cref{sec:Applications}) so that the interested readers can have a head start on the exploration of graph analytics.

In addition to the static graph embedding methods, we also discuss the emerging deep learning-based {\em dynamic} graph embedding methods, which can be particularly computational expensive due to the evolution of the network (e.g., describing a biological system). Hence, careful consideration is required to select the proper method from a host of different methods, which we summarize in \Cref{tab:dynGEM} in the Appendix.
%, where we also give more details for the more advanced readers on the mathematical problem setting.

Finally, for application highlights we present results from four different  domains. In the first one, we consider a social network represented by the karate club benchmark for which we employ the DeepWalk for community detection. In the second one, we consider a directed and attributed citation network for the CORA-ML dataset, and we employ the stochastic embedding method G2G for node classification. In the third application, we analyze multi-modality functional brain networks (fMRI and MEG) using MG2G, which is an extension of G2G, for cognitive training effect evaluation and for predicting AD progression. Finally, in the last example we present two different applications of deterministic (LINE method) and stochastic (set2Gauss) for genomic networks. Taken together, the results presented in \cref{sec:Applications} demonstrate the effectiveness of the graph embedding methods in transforming high-dimensional heterogeneous networks into low-dimensional compact vectors or probability density functions that can lead to great efficiencies in processing downstream tasks depending on the particular application. In addition, using a distributed parallel architecture for training embeddings on very large-scale graphs, graph embedding at scale can be achieved for industrial systems, as demonstrated in \cite{bruss2019graph} for the aforementioned embedding algorithm SkipGram~\cite{mikolov2013efficient} on the massive open-source Friendster graph~\cite{snapnets} with 68 million nodes and 2.5 billion edges.

%%%%%combine appendix for arXiv submission %%%%%
\appendix
\renewcommand{\thetable}{\Alph{section}\arabic{table}}
\section{Similarity measures used in the latent space}
\label{sec:similarity_latent_space}
We summarize some basic similarity measures commonly used in the vector point-based graph embedding approaches and Gaussian distribution-based graph embedding approaches in \cref{tab:similarity_embedding_space}.
%%%%%%%%%%%%%%%%%%%%%%
\begin{table}[htbp]
{\footnotesize
\caption{\textbf{Common similarity measures for the graph node embeddings in the latent space.}}\label{tab:similarity_embedding_space}
\begin{center}
    \begin{tabular}{|l|l|}\hline
         Node representations & Similarity measures ($M$)\\\hline
         Point vectors & 1) Dot product: $z_j^Tz_i$ \\
         &2) Cosine similarity: $\frac{z_i\cdot z_j}{||z_i||\times||z_j||}$\\
         &3) Euclidean distance: $ ||z_i-z_j||$\\\hline
         Gaussian distributions & 1) Expected likelihood: $EL(P_i,P_j)$ (see equation \cref{eq:EL-energy-func})\\
         & 2) KL-divergence: $D_{KL}(P_j||P_i))$ (see equation \cref{eq:kl-energy-func})\\
         & 3) Wasserstein distance: $W_2(P_i; P_j)$(see equation \cref{eq:W2-distance}) \\\hline
    \end{tabular}
\end{center}
}
\end{table}
%%%%%%%%%%%%%%%%%%%%%%

\section{Summary of Graph Gaussian embedding methods}
\label{sec:Gauss_emb_methods}
\setcounter{table}{0}
We summarize some Graph Gaussian embedding methods in \Cref{tab:gauss_emb_surveytable}. The complexity of G2G method and DVNE method 
is  $O(N)$ and $O(T \times N \times (d\times S +  S\times L + L))$, respectively, where $T$ is the number of iterations, $N$ is the number of nodes in the graph, $d$ is the average degree of all nodes, $S$ is the size of hidden layer of the encoder and decoder, and $L$ is embedding size.
%%%
\begin{table}[htbp]
{\footnotesize
\caption{\textbf{Graph Gaussian embedding methods surveyed in this study.} KG2E is for directed knowledge graphs; DVNE is for undirected, non-attributed Graphs; \textit{Graph2Gauss} is for attributed/non-attributed/directed/undirected graphs; MG2G is for multiple undirected/attributed graphs. $T$ refers to the number of iterations.}\label{tab:gauss_emb_surveytable}
\begin{center}
    \begin{tabular}{|l|l|l|l|l|}\hline
    % row 1
    \multirow{2}{*}{Method} & Energy & \multirow{2}{*}{Loss function} & \multirow{2}{*}{Sampling} & \multirow{2}{*}{Optimization} \\
    & function &&&\\\hline
    % row 2
    KG2E~\cite{he2015learning} & KL/EL & margin-based & Unif/bern& SGD \\
    % row 3
    G2G~\cite{bojchevski2017deep} & KL & square exponential& Anchor-based & Adam \\
    & & &sampling&\\
    % row 4
    DVNE~\cite{zhu2018deep} & W2 & hybrid loss & -- & SGD($T \le 50,$\\
    &&&& $converge)$ \\
    % row 5
    MG2G~\cite{xu2019gaussian} & KL & square exponential & Edge weight-based&\\
    &&&k-hop&\\
    &&&neighborhood&\\\hline
    \end{tabular}
\end{center}
}
\end{table}
%%%%%%%%%%%%%%%%%%%%%%%%%

\section{Comparisons of random walk-based graph embedding methods}\label{sec:rw_emb_methods}
\setcounter{table}{0}
In \Cref{tab:RW_compare}, we compared three random walk-based graph node embedding approaches (DeepWalk, LINE and node2vec) based on five key properties: node neighborhood sampling techniques, proximity, learning mode, optimization and labeled data.

%%%%%%%%%%%%%%%%%%%%%%%%%

\begin{table}[htbp]
{\footnotesize
 \caption{\textbf{Comparison of random walk-based graph embedding methods.} Among the three graph embedding approaches, ``DeepWalk'' supports ``online'' embedding learning since it only preserves local network structure information without considering global network structure preservation during the entire embedding procedure. Moreover, all of these three methods use ``shallow networks''. BFS stands for breadth-first sampling, DFS stands for depth-first sampling.}
\label{tab:RW_compare}
\begin{center}
    \begin{tabular}{|l|l|l|l|l|l|}\hline
        % item row
        \multirow{2}{*}{Method} & Neighborhood & 
        \multirow{2}{*}{Proximity} & \multirow{2}{*}{Mode} &
        \multirow{2}{*}{Optimization} & labeled\\
        &sampling&&&&data \\\hline
        % row 1
        \multirow{2}{*}{DeepWalk~\cite{perozzi2014deepwalk}} & Truncated random & \multirow{2}{*}{1st-order} & \multirow{2}{*}{online} & \multirow{2}{*}{hierarchical softmax} & \multirow{2}{*}{No}\\
        & walk &&&&\\
        % row 2
        \multirow{2}{*}{LINE~\cite{tang2015line}}& \multirow{2}{*}{BFS} & 1st-order/ &\multirow{2}{*}{-}& negative sampling & No\\
        && 2nd-order & & & \\
        % row 3
        node2vec~\cite{grover2016node2vec} & BFS+DFS & 2nd-order &-& negative sampling & Yes\\\hline  
    \end{tabular}
\end{center}
}
\end{table}
\section{Overview of recent dynamic graph embedding methods}
\label{sec:dyn_GEM}
\setcounter{table}{0}
In the following \cref{tab:dynGEM}, we present references to five main classes of dynamic graph embedding models.
%%%%%%%%%%%%%%%%%%%%%%%%%%%%
\begin{table}[htbp]
{\footnotesize
\caption{\textbf{Dynamic graph embedding methods surveyed in this study.}}
\label{tab:dynGEM}
\begin{center}
    \begin{tabular}{|l|l|l|}\hline
        Type & Model & code\\\hline
        \multirow{3}{2cm}{Matrix factorization -based}& DANE\cite{li2017attributed} &https://github.com/yyyouc/DANE \\
        & TIMERS~\cite{zhang2018timers} & https://github.com/ZW-ZHANG/TIMERS\\
               & DHPE~\cite{zhu2018high} & https://github.com/BingSu12/HDPE\\\hline
        
        \multirow{3}{2cm}{SkipGram-based}& DNE~\cite{DNE2018} & --\\
                & CTDNE~\cite{nguyen2018dynamic} &https://github.com/LogicJake/CTDNE\\        &dynnode2vec~\cite{mahdavi2018dynnode2vec}&--\\ \hline
        
        \multirow{2}{2cm}{Autoencoder -based, RNN}& DynGEM~\cite{goyal2018dyngem} &https://github.com/paulpjoby/DynGEM \\
         & Dyngraph2vec~\cite{goyal2020dyngraph2vec}&https://github.com/palash1992/DynamicGEM\\
         & NetWalk~\cite{yu2018netwalk} &https://github.com/chengw07/NetWalk\\\hline
        
        \multirow{2}{2cm}{GCN-based}& DyRep~\cite{trivedi2018dyrep} &--  \\
        &TDGNN~\cite{qu2020continuous} & https://github.com/stefanosantaris/TDGNN\\
        &DySAT~\cite{sankar2020dysat}  & https://github.com/aravindsankar28/DySAT\\
        & EvolveGCN~\cite{pareja2020evolvegcn}& https://github.com/IBM/EvolveGCN\\\hline
        \multirow{2}{2cm}{GAN-based}& DynGraphGAN~\cite{xiong2019dyngraphgan}  & --\\
        & DynGAN~\cite{maheshwaridyngan} &--\\\hline
    \end{tabular}
\end{center}
}
\end{table}
%%%%%%%%%%%%%%%%%%%%%%%%%%%%
%%
\section{Implementation Details}
\label{sec:implement}
In this section, we introduce specific implementations for two graph embedding based on node2vec \cite{grover2016node2vec} and G2G \cite{bojchevski2017deep} methods. We use two different experimental datasets, which are, respectively, an undirected graph (karate club dataset \cite{zachary1977information}) without attributes and a directed graph (CORA-ML dataset \cite{bojchevski2017deep}) with extra attributes. The code for these two experiments is written in Python language, incorporating {\em NetworkX} (network analysis library), {\em Gensim} (natural language processing library), {\em scikit-learn} (machine learning library) and {\em TensorFlow} (deep learning library).

\subsection{Karate club network embedding based on {\em node2vec}}
\label{ipl_kc}
To perform graph embedding using the \textit{node2vec} method for the undirected and non-attributed karate club network ($G1 = (V,E)$), we proceed in six steps: 

% ndoe2vec: 3 phases
1) Read the original graph ($G1$) from the edge list file saved in $input\_dir$, see Line \ref{ln:read_graph} in \cref{alg:karate_emb}. Additionally, the karate network dataset (34 nodes, 77 undirected edges, 2 communities) can also be imported using other packages, e.g., DGL package \cite{wang2019deep} and TORCH\_GEOMETRIC library~\cite{Fey/Lenssen/2019}. 
2) Create a graph instance object ($G$) for the \textit{Graph} operation class, initialize the graph, and node neighborhood sampling parameters (return factor $p$ and in-out factor $q$) with preset default values (p = q = 1), respectively, for the BFS and DFS strategies.
3) Compute the transition probability for simulating random walks for nodes in the graph (see Line \ref{ln:preprocessing} in \cref{alg:karate_emb}); the detailed preprocessing procedure can be found in \cite{grover2016node2vec}.
4) Simulate random walks for all nodes in the input graph in terms of the preset random walk hyper-parameters ($num\_walks$ and $walk\_length$) and the computed transition probabilities obtained in 3), see Line \ref{ln:RW} in \cref{alg:karate_emb}. 
5) Train the neural network-based ``word2vec'' (SkipGram) model on the training dataset (i.e., generated walks or called ``node sequences'') obtained from step 4) with a stochastic gradient descent (SGD) optimizer in conjunction with {\em negative sampling} policy in an efficient and unsupervised manner (see Line \ref{ln:train_node2vec} in \cref{alg:karate_emb}), where the context size $c$ is a preset factor for sampling node context for training.
6) Visualize graph embedding vectors in 2D space by the t-SNE tool in the {\em scikit-learn} library (see Line \ref{ln:visual_emb} in \cref{alg:karate_emb}); all nodes can be projected into 2D space as \textit{points}.

The detailed python implementation procedure and functions for the karate club graph embedding using \textit{node2vec} approach are shown in \cref{alg:karate_emb}.
\begin{algorithm}[H]
\caption{Karate club network embedding based on the \textit{node2vec}}\label{alg:karate_emb}
\begin{algorithmic}[1]
\STATE{\textbf{Procedure}: KARATE\_EMBEDDING($input\_dir, num\_walks, walk\_length, p, q,$\\$L, c, num\_iter$)}
\STATE{$G1$ = read\_graph($input\_dir$)}\COMMENT{read undirected karate graph}\label{ln:read_graph}
\IF{$G1$ is not None}
    \STATE{$G2$ = preprocess\_transition\_probs($G1$, $p$, $q$)}\COMMENT{preprocessing the graph}\label{ln:preprocessing}
    \STATE{$walks$ = simulate\_walks($G2$, $num\_walks$, $walk\_length$)}\label{ln:RW}\COMMENT{generate walks}
    \STATE{$embeddings$ = learn\_embeddings($walks, L, c, num\_iter$)}\label{ln:train_node2vec}
    \STATE{vis\_vecEmb($embeddings$)} \COMMENT{visualize embeddings}\label{ln:visual_emb}
\ENDIF
\\\hrulefill\\
% \Function{simulate\_walks}{$G2, num\_walks, walk\_length$}
\STATE{\textbf{Function}: SIMULATE\_WALKS ($G2, num\_walks, walk\_length$)}
    \STATE{Initialize $walks$ to Empty}
    \FOR{$iter \gets 0$ to $num\_walks$}
        \FOR{all nodes in $G2$}
            \STATE{$walk$ = node2vec\_Walk($G2, u, walk\_length$)}\COMMENT{$u$ is the starting node} 
            \STATE{Append $walk$ to $walks$}
        \ENDFOR
    \ENDFOR
    \STATE{\textbf{return} $walks$}
\STATE{\textbf{End Function}}
\\\hrulefill\\
\STATE{\textbf{Function}: LEARN\_EMBEDDINGS($walks, L, c, num\_iter$)}
\IF{$walks$ is not Empty}
    \STATE{Map node indices as string type in $walks$}
    \STATE{model = Word2Vec($walks, L, num\_iter, c$)}\COMMENT{$c$ is the context size}
    \STATE{$embeddings$ = (model.wv.vectors)} \COMMENT{karate graph embeddings}
    \STATE{model.wv.save\_word2vec\_format($output$)}\COMMENT{save embeddings}
    \STATE{\textbf{return} $embeddings$}
\ENDIF
\STATE{\textbf{End Function}}
\\\hrulefill\\
\STATE{\textbf{Function}: VIS\_VECEMB($embeddings, L$)}\label{func:vis_emb}
\IF{$L$ is larger than 2}
     \STATE{$t\_emb$ = tsne\_project($embeddings, dim = 2$)}\COMMENT{transform embeddings into 2D space}
\ELSIF{$L$ equals to 2}
    \STATE{$t\_emb = embeddings$}
\ENDIF
\STATE{Plot the embedded nodes' positions $t\_emb$}\COMMENT{see \cref{fig:social_net} (B)}
\STATE{\textbf{End Function}}
\end{algorithmic}
\end{algorithm}
\subsection{CORA-ML network embedding based on stochastic graph embedding}\label{ipl_cml}
The CORA-ML dataset from the G2G paper \cite{bojchevski2017deep} represents a citation graph ($G = (A,X)$), where $A\in \mathbb{R}^{N\times N}$ is the adjacency matrix, and $X\in \mathbb{R}^{N\times D}$ is the attributed matrix. $G$ contains $N$ = 2995 nodes, $E$ = 8416 \textit{directed} edges and $D$ = 2879 dimensional \textit{attribute vector} for every node. In order to learn stochastic graph embeddings for this directed and attributed CORA-ML graph, each node is projected into a multivariate Gaussian distribution ($\mathbb{N} = (\mu_i, \Sigma_i)$) with mean ($\mu_i \in \mathbb{R}^{L/2}$) and covariance matrix ($\Sigma_i \in \mathbb{R}^{L/2\times L/2}$). This embedding process mainly follows five steps (see detailed pseudo code description in \cref{alg:coraml_emb}):

% load_dataset('%filepath%')
1) Load the input graph from the file path (``filepath''), and obtain the adjacency matrix ($A$), attribute matrix ($X$) and node labels ($z$) for the subsequent stochastic graph embedding, see Line \ref{ln:load_data} in \cref{alg:coraml_emb}.
% __dataset_generator(hops, scale_terms)<--get_hops() // important
2) Sample node triplets based on ``k-hop neighborhoods'' ($K$); each node triplet comprises an anchor node and the corresponding \textit{positive} and \textit{negative} nodes w.r.t. the anchor node.
% train_val_test_split_adjacency()
3) In order to evaluate the G2G model and demonstrate the effectiveness of the learned embeddings through performing ``link prediction'' task on a ``validation set'', graph edges in $A$ are split into three subsets (training, validation and test sets) based on the input ratios of validation and test dataset ($p\_val = 0.10, p\_test = 0.05$). Moreover, equal number of non-edges are sampled for validation and test.
% g2g.train()--square-exponential loss
4) Train the deep encoder-based G2G model with unsupervised manner, and minimize the square-exponential loss function (see equation in \Cref{eq:square-exp-loss}) with the Adam optimizer (learning rate = 1e-3) provided in TensorFlow ($\geq 1.4$); the output Gaussian embeddings consist of mean and variance vectors (see Line \ref{ln:emb_vectors} of \cref{alg:coraml_emb})
% vis\_embedding(embeddings)
5) Visualize the stochastic graph embeddings into 2D space using the t-SNE tool in {\em scikit-learn}; the $vis\_GaussEmb$ function (described in Line \ref{func:vis_gaussemb} in \cref{alg:coraml_emb}) was implemented to project every node's Gaussian embeddings (both \textit{mean} and \textit{variance} vectors) into 2D Euclidean space as a ``probability ellipse'' (see \cref{fig:uncertainty_benefits}(B)). Specifically, the centroid of each ellipse represents the \textit{mean} projection result, and the contour represents the predicted uncertainty of the node embedding by \textit{sigma} projection using t-SNE.
% score_node_classification(mu, z, n_repeat=10, norm=True)
6) Apply the learned graph embeddings (``mean'' or ``variance'' vectors) for downstream tasks, e.g., node classification (see the example of implementation in Line \ref{fun:nc} in \cref{alg:coraml_emb}).

%%%%%%%%%%%%%%%%%%%%
\begin{algorithm}
\caption{CORA-ML stochastic embedding based on \textit{G2G}}
\label{alg:coraml_emb}
\begin{algorithmic}[1]
\STATE{\textbf{Procedure}: GAUSSIAN\_EMBEDDING (data path $filepath$, hop $k$, embedding size $L$)}
\STATE{[A, X, z] = load\_dataset($filepath$)}\COMMENT{load graph data}\label{ln:load_data}
\WHILE{A is not Empty}
    \STATE{$g2g$ = Graph2Gauss($A, X, L, k, p\_val, p\_test, p\_nodes$)}
    \STATE{$sess$ = $g2g$.train()} \COMMENT{train the G2G model}
    \STATE{$mu, sigma$ = $sess$.run([$g2g.mu, g2g.sigma$])}
    \COMMENT{output Gaussian embeddings}\label{ln:emb_vectors}
    \STATE{$test\_auc, test\_ap$ = link\_pred($test\_label,  test\_edges$)}\label{ln:lp}
    \STATE{$f1\_micro, f1\_macro$ = node\_clf($mu, z, n\_repeat = 10, norm=True$)}\label{ln:nc}
    \STATE{vis\_GaussEmb($mu, sigma, L$)}\label{ln:vis_gauss_emb}
\ENDWHILE
%%%%%%%%%%func: link prediction%%%%%%
\\\hrulefill\\
\STATE{\textbf{Function}: LINK\_PRED(test\_labels, test\_edges)}\label{fun:lp}
\WHILE{$test\_edges$ is not Null}
\STATE{$scores$ = energy\_kl($test\_edges$)}\label{ln:kl_test_links}
\STATE{$auc$ = roc\_auc\_score($test\_labels, scores$)}
\STATE{$ap$ = average\_precision\_score($test\_labels, scores$)}
\ENDWHILE
\RETURN $auc$, $ap$
\STATE{\textbf{End Function}}
%%%%%%%%%%func: node classification%%%%%%
\\\hrulefill\\
\STATE{\textbf{Function}: NODE\_CLF($mu, z, n\_repeat = 10, norm = True$)}\label{fun:nc}
\STATE{lrcv = LogisticRegressionCV()}
\IF{$norm$ is TRUE}
    \STATE{Normalize the feature matrix $mu$}
\ENDIF
\STATE{Initialize $F1\_score$ to Empty}
\FOR{$seed \gets 0$ to $n\_repeat$}
    \STATE{Shuffle and split feature matrix ($mu$) and labels ($z$)}
    \STATE{lrcv.fit($x\_train, y\_train$)}
    \STATE{$predict$ = lrcv.predict($x\_test$)}
    \STATE{$f1\_micro$ = f1\_score($y\_test, predict, average = 'micro'$)}
    \STATE{$f1\_macro$ = f1\_score($y\_test, predict, average = 'macro'$)}
    \STATE{Append $(f1\_micro, f1\_macro)$ to $F1\_score$}
\ENDFOR
\RETURN{Mean $F1\_score$}
\STATE{\textbf{End Function}}
%%%%%%%%%
\\\hrulefill\\
\STATE{\textbf{Function}: VIS\_GAUSSEMB($mu, sigma, L$)}\label{func:vis_gaussemb}
\IF{$L$ is larger than 2}
     \STATE{$t\_mu, t\_sigma$ = tsne\_project($mu,sigma$)}\COMMENT{transform into 2D space}
\ELSIF{$L$ equals to 2}
    \STATE{$t\_mu = mu$, $t\_sigma = sigma$}
\ENDIF
\STATE{Plot the embedded positions ($t\_mu$) for all vertices} \label{ln:plot_mean}
\STATE{Plot the embedded uncertainty ($t\_sigma$) for all vertices}\label{ln:plot_variance}
\STATE{\textbf{End Function}}

\end{algorithmic}
\end{algorithm}
%%%%%%%%%%%%%%%%%%%%

\clearpage

% \bibliographystyle{siamplain}
% \bibliography{references}

\begin{thebibliography}{10}

\bibitem{ashoor2020graph}
{\sc H.~Ashoor, X.~Chen, W.~Rosikiewicz, J.~Wang, A.~Cheng, P.~Wang, Y.~Ruan,
  and S.~Li}, {\em Graph embedding and unsupervised learning predict genomic
  sub-compartments from hic chromatin interaction data}, Nature communications,
  11 (2020), pp.~1--11.

\bibitem{belkin2002laplacian}
{\sc M.~Belkin and P.~Niyogi}, {\em Laplacian eigenmaps and spectral techniques
  for embedding and clustering}, in Advances in neural information processing
  systems, 2002, pp.~585--591.

\bibitem{bertasius2017convolutional}
{\sc G.~Bertasius, L.~Torresani, S.~X. Yu, and J.~Shi}, {\em Convolutional
  random walk networks for semantic image segmentation}, in Proceedings of the
  IEEE Conference on Computer Vision and Pattern Recognition, 2017,
  pp.~858--866.

\bibitem{bojchevski2017deep}
{\sc A.~Bojchevski and S.~G{\"u}nnemann}, {\em Deep gaussian embedding of
  graphs: Unsupervised inductive learning via ranking}, in ICLR, 2018.

\bibitem{bollacker2008freebase}
{\sc K.~Bollacker, C.~Evans, P.~Paritosh, T.~Sturge, and J.~Taylor}, {\em
  Freebase: a collaboratively created graph database for structuring human
  knowledge}, in Proceedings of the 2008 ACM SIGMOD international conference on
  Management of data, 2008, pp.~1247--1250.

\bibitem{brin2017anatomy}
{\sc S.~Brin and L.~Page}, {\em The anatomy of a large-scale hypertextual web
  search engine. 1998}, in Proceedings of the Seventh World Wide Web
  Conference, 2017.

\bibitem{bruss2019deeptrax}
{\sc C.~B. Bruss, A.~Khazane, J.~Rider, R.~Serpe, A.~Gogoglou, and K.~E.
  Hines}, {\em Deeptrax: Embedding graphs of financial transactions}, arXiv
  preprint arXiv:1907.07225,  (2019).

\bibitem{bruss2019graph}
{\sc C.~B. Bruss, A.~Khazane, J.~Rider, R.~Serpe, S.~Nagrecha, and K.~E.
  Hines}, {\em Graph embeddings at scale}, arXiv preprint arXiv:1907.01705,
  (2019).

\bibitem{cai2018comprehensive}
{\sc H.~Cai, V.~W. Zheng, and K.~C.-C. Chang}, {\em A comprehensive survey of
  graph embedding: Problems, techniques, and applications}, IEEE Transactions
  on Knowledge and Data Engineering, 30 (2018), pp.~1616--1637.

\bibitem{cao2015grarep}
{\sc S.~Cao, W.~Lu, and Q.~Xu}, {\em Grarep: Learning graph representations
  with global structural information}, in Proceedings of the 24th ACM
  international on conference on information and knowledge management, 2015,
  pp.~891--900.

\bibitem{cui2018survey}
{\sc P.~Cui, X.~Wang, J.~Pei, and W.~Zhu}, {\em A survey on network embedding},
  IEEE Transactions on Knowledge and Data Engineering, 31 (2018), pp.~833--852.

\bibitem{ding2019deep}
{\sc K.~Ding, J.~Li, R.~Bhanushali, and H.~Liu}, {\em Deep anomaly detection on
  attributed networks}, in Proceedings of the 2019 SIAM International
  Conference on Data Mining, SIAM, 2019, pp.~594--602.

\bibitem{Fey/Lenssen/2019}
{\sc M.~Fey and J.~E. Lenssen}, {\em Fast graph representation learning with
  {PyTorch Geometric}}, in ICLR Workshop on Representation Learning on Graphs
  and Manifolds, 2019.

\bibitem{goyal2020dyngraph2vec}
{\sc P.~Goyal, S.~R. Chhetri, and A.~Canedo}, {\em dyngraph2vec: Capturing
  network dynamics using dynamic graph representation learning},
  Knowledge-Based Systems, 187 (2020), p.~104816.

\bibitem{goyal2018dyngem}
{\sc P.~Goyal, N.~Kamra, X.~He, and Y.~Liu}, {\em Dyngem: Deep embedding method
  for dynamic graphs}, arXiv preprint arXiv:1805.11273,  (2018).

\bibitem{grover2016node2vec}
{\sc A.~Grover and J.~Leskovec}, {\em node2vec: Scalable feature learning for
  networks}, in Proceedings of the 22nd ACM SIGKDD international conference on
  Knowledge discovery and data mining, 2016, pp.~855--864.

\bibitem{he2015learning}
{\sc S.~He, K.~Liu, G.~Ji, and J.~Zhao}, {\em Learning to represent knowledge
  graphs with gaussian embedding}, in Proceedings of the 24th ACM International
  on Conference on Information and Knowledge Management, 2015, pp.~623--632.

\bibitem{hinton2003stochastic}
{\sc G.~E. Hinton and S.~T. Roweis}, {\em Stochastic neighbor embedding}, in
  Advances in neural information processing systems, 2003, pp.~857--864.

\bibitem{kashyap2018protein}
{\sc S.~Kashyap, S.~Kumar, V.~Agarwal, D.~P. Misra, S.~R. Phadke, and
  A.~Kapoor}, {\em Protein protein interaction network analysis of
  differentially expressed genes to understand involved biological processes in
  coronary artery disease and its different severity}, Gene Reports, 12 (2018),
  pp.~50--60.

\bibitem{kurose1992computing}
{\sc J.~Kurose}, {\em On computing per-session performance bounds in high-speed
  multi-hop computer networks}, ACM SIGMETRICS Performance Evaluation Review,
  20 (1992), pp.~128--139.

\bibitem{lecun2006tutorial}
{\sc Y.~LeCun, S.~Chopra, R.~Hadsell, M.~Ranzato, and F.~Huang}, {\em A
  tutorial on energy-based learning}, Predicting structured data, 1 (2006).

\bibitem{snapnets}
{\sc J.~Leskovec and A.~Krevl}, {\em {SNAP Datasets}: {Stanford} large network
  dataset collection}.
\newblock \url{http://snap.stanford.edu/data}, June 2014.

\bibitem{leskovec2012learning}
{\sc J.~Leskovec and J.~J. Mcauley}, {\em Learning to discover social circles
  in ego networks}, in Advances in neural information processing systems, 2012,
  pp.~539--547.

\bibitem{li2017attributed}
{\sc J.~Li, H.~Dani, X.~Hu, J.~Tang, Y.~Chang, and H.~Liu}, {\em Attributed
  network embedding for learning in a dynamic environment}, in Proceedings of
  the 2017 ACM on Conference on Information and Knowledge Management, 2017,
  pp.~387--396.

\bibitem{liu2019chemi}
{\sc K.~Liu, X.~Sun, L.~Jia, J.~Ma, H.~Xing, J.~Wu, H.~Gao, Y.~Sun,
  F.~Boulnois, and J.~Fan}, {\em Chemi-net: a molecular graph convolutional
  network for accurate drug property prediction}, International journal of
  molecular sciences, 20 (2019), p.~3389.

\bibitem{mahdavi2018dynnode2vec}
{\sc S.~Mahdavi, S.~Khoshraftar, and A.~An}, {\em dynnode2vec: Scalable dynamic
  network embedding}, in 2018 IEEE International Conference on Big Data (Big
  Data), IEEE, 2018, pp.~3762--3765.

\bibitem{maheshwaridyngan}
{\sc A.~Maheshwari, A.~Goyal, M.~K. Hanawal, and G.~Ramakrishnan}, {\em Dyngan:
  Generative adversarial networks for dynamic network embedding}, in Graph
  Representation Learning Workshop at NeurIPS, 2019.

\bibitem{mcauley2012image}
{\sc J.~McAuley and J.~Leskovec}, {\em Image labeling on a network: using
  social-network metadata for image classification}, in European conference on
  computer vision, Springer, 2012, pp.~828--841.

\bibitem{cora2000}
{\sc A.~K. McCallum, K.~Nigam, J.~Rennie, and K.~Seymore}, {\em Automating the
  construction of internet portals with machine learning}, Information
  Retrieval, 3 (2000), pp.~127--163.

\bibitem{mheich2020brain}
{\sc A.~Mheich, F.~Wendling, and M.~Hassan}, {\em Brain network similarity:
  methods and applications}, Network Neuroscience, 4 (2020), pp.~507--527.

\bibitem{mikolov2013efficient}
{\sc T.~Mikolov, K.~Chen, G.~Corrado, and J.~Dean}, {\em Efficient estimation
  of word representations in vector space}, arXiv preprint arXiv:1301.3781,
  (2013).

\bibitem{miller1998wordnet}
{\sc G.~A. Miller}, {\em WordNet: An electronic lexical database}, MIT press,
  1998.

\bibitem{nerurkar2019survey}
{\sc P.~Nerurkar, M.~Chandane, and S.~Bhirud}, {\em Survey of network embedding
  techniques for social networks}, Turkish Journal of Electrical Engineering \&
  Computer Sciences, 27 (2019), pp.~4768--4782.

\bibitem{nguyen2018dynamic}
{\sc G.~H. Nguyen, J.~B. Lee, R.~A. Rossi, N.~K. Ahmed, E.~Koh, and S.~Kim},
  {\em Dynamic network embeddings: From random walks to temporal random walks},
  in 2018 IEEE International Conference on Big Data (Big Data), IEEE, 2018,
  pp.~1085--1092.

\bibitem{nikolakopoulos2019recwalk}
{\sc A.~N. Nikolakopoulos and G.~Karypis}, {\em Recwalk: Nearly uncoupled
  random walks for top-n recommendation}, in Proceedings of the Twelfth ACM
  International Conference on Web Search and Data Mining, 2019, pp.~150--158.

\bibitem{okuda2019community}
{\sc M.~Okuda, S.~Satoh, Y.~Sato, and Y.~Kidawara}, {\em Community detection
  using restrained random-walk similarity}, IEEE Transactions on Pattern
  Analysis and Machine Intelligence,  (2019).

\bibitem{ou2016asymmetric}
{\sc M.~Ou, P.~Cui, J.~Pei, Z.~Zhang, and W.~Zhu}, {\em Asymmetric transitivity
  preserving graph embedding}, in Proceedings of the 22nd ACM SIGKDD
  international conference on Knowledge discovery and data mining, 2016,
  pp.~1105--1114.

\bibitem{pareja2020evolvegcn}
{\sc A.~Pareja, G.~Domeniconi, J.~Chen, T.~Ma, T.~Suzumura, H.~Kanezashi,
  T.~Kaler, T.~B. Schardl, and C.~E. Leiserson}, {\em Evolvegcn: Evolving graph
  convolutional networks for dynamic graphs.}, in AAAI, 2020, pp.~5363--5370.

\bibitem{perozzi2014deepwalk}
{\sc B.~Perozzi, R.~Al-Rfou, and S.~Skiena}, {\em Deepwalk: Online learning of
  social representations}, in Proceedings of the 20th ACM SIGKDD international
  conference on Knowledge discovery and data mining, 2014, pp.~701--710.

\bibitem{qu2020continuous}
{\sc L.~Qu, H.~Zhu, Q.~Duan, and Y.~Shi}, {\em Continuous-time link prediction
  via temporal dependent graph neural network}, in Proceedings of The Web
  Conference 2020, 2020, pp.~3026--3032.

\bibitem{ringner2008principal}
{\sc M.~Ringn{\'e}r}, {\em What is principal component analysis?}, Nature
  biotechnology, 26 (2008), pp.~303--304.

\bibitem{rosenthal2018mapping}
{\sc G.~Rosenthal, F.~V{\'a}{\v{s}}a, A.~Griffa, P.~Hagmann, E.~Amico,
  J.~Go{\~n}i, G.~Avidan, and O.~Sporns}, {\em Mapping higher-order relations
  between brain structure and function with embedded vector representations of
  connectomes}, Nature communications, 9 (2018), pp.~1--12.

\bibitem{roweis2000nonlinear}
{\sc S.~T. Roweis and L.~K. Saul}, {\em Nonlinear dimensionality reduction by
  locally linear embedding}, science, 290 (2000), pp.~2323--2326.

\bibitem{sankar2020dysat}
{\sc A.~Sankar, Y.~Wu, L.~Gou, W.~Zhang, and H.~Yang}, {\em Dysat: Deep neural
  representation learning on dynamic graphs via self-attention networks}, in
  Proceedings of the 13th International Conference on Web Search and Data
  Mining, 2020, pp.~519--527.

\bibitem{DNE2018}
{\sc X.~Shen, S.~Pan, W.~Liu, Y.-S. Ong, and Q.-S. Sun}, {\em Discrete network
  embedding}, in Proceedings of the 27th International Joint Conference on
  Artificial Intelligence, 2018, pp.~3549--3555.

\bibitem{su2020network}
{\sc C.~Su, J.~Tong, Y.~Zhu, P.~Cui, and F.~Wang}, {\em Network embedding in
  biomedical data science}, Briefings in bioinformatics, 21 (2020),
  pp.~182--197.

\bibitem{tan2019deep}
{\sc Q.~Tan, N.~Liu, and X.~Hu}, {\em Deep representation learning for social
  network analysis}, Frontiers in Big Data, 2 (2019), p.~2.

\bibitem{tang2015line}
{\sc J.~Tang, M.~Qu, M.~Wang, M.~Zhang, J.~Yan, and Q.~Mei}, {\em Line:
  Large-scale information network embedding}, in Proceedings of the 24th
  international conference on world wide web, 2015, pp.~1067--1077.

\bibitem{tenenbaum2000global}
{\sc J.~B. Tenenbaum, V.~De~Silva, and J.~C. Langford}, {\em A global geometric
  framework for nonlinear dimensionality reduction}, science, 290 (2000),
  pp.~2319--2323.

\bibitem{trivedi2018dyrep}
{\sc R.~Trivedi, M.~Farajtabar, P.~Biswal, and H.~Zha}, {\em Dyrep: Learning
  representations over dynamic graphs}, in International Conference on Learning
  Representations, 2018.

\bibitem{vilnis2014word}
{\sc L.~Vilnis and A.~McCallum}, {\em Word representations via gaussian
  embedding}, arXiv preprint arXiv:1412.6623,  (2014).

\bibitem{wang2016structural}
{\sc D.~Wang, P.~Cui, and W.~Zhu}, {\em Structural deep network embedding}, in
  Proceedings of the 22nd ACM SIGKDD international conference on Knowledge
  discovery and data mining, 2016, pp.~1225--1234.

\bibitem{wang2019deep}
{\sc M.~Wang, L.~Yu, D.~Zheng, Q.~Gan, Y.~Gai, Z.~Ye, M.~Li, J.~Zhou, Q.~Huang,
  C.~Ma, et~al.}, {\em Deep graph library: Towards efficient and scalable deep
  learning on graphs}, arXiv preprint arXiv:1909.01315,  (2019).

\bibitem{wang2020gaussian}
{\sc S.~Wang, E.~R. Flynn, and R.~B. Altman}, {\em Gaussian embedding for
  large-scale gene set analysis}, Nature Machine Intelligence, 2 (2020),
  pp.~387--395.

\bibitem{wang2019user}
{\sc Y.~Wang, C.~Feng, L.~Chen, H.~Yin, C.~Guo, and Y.~Chu}, {\em User identity
  linkage across social networks via linked heterogeneous network embedding},
  World Wide Web, 22 (2019), pp.~2611--2632.

\bibitem{wang2014knowledge}
{\sc Z.~Wang, J.~Zhang, J.~Feng, and Z.~Chen}, {\em Knowledge graph embedding
  by translating on hyperplanes.}, in AAAI, vol.~14, 2014, pp.~1112--1119.

\bibitem{west2016recommendation}
{\sc J.~D. West, I.~Wesley-Smith, and C.~T. Bergstrom}, {\em A recommendation
  system based on hierarchical clustering of an article-level citation
  network}, IEEE Transactions on Big Data, 2 (2016), pp.~113--123.

\bibitem{xiong2019dyngraphgan}
{\sc Y.~Xiong, Y.~Zhang, H.~Fu, W.~Wang, Y.~Zhu, and S.~Y. Philip}, {\em
  Dyngraphgan: Dynamic graph embedding via generative adversarial networks}, in
  International Conference on Database Systems for Advanced Applications,
  Springer, 2019, pp.~536--552.

\bibitem{xu2017deep}
{\sc M.~Xu, D.~P. Papageorgiou, S.~Z. Abidi, M.~Dao, H.~Zhao, and G.~E.
  Karniadakis}, {\em A deep convolutional neural network for classification of
  red blood cells in sickle cell anemia}, PLoS computational biology, 13
  (2017), p.~e1005746.

\bibitem{xu2020graph}
{\sc M.~Xu, D.~L. Sanz, P.~Garces, F.~Maestu, Q.~Li, and D.~Pantazis}, {\em A
  graph gaussian embedding method for predicting alzheimer's disease
  progression with meg brain networks}, arXiv preprint arXiv:2005.05784,
  (2020).

\bibitem{xu2019gaussian}
{\sc M.~Xu, Z.~Wang, H.~Zhang, D.~Pantazis, H.~Wang, and Q.~Li}, {\em Gaussian
  embedding-based functional brain connectomic analysis for amnestic mild
  cognitive impairment patients with cognitive training}, bioRxiv,  (2019),
  p.~779744.

\bibitem{xu2020new}
{\sc M.~Xu, Z.~Wang, H.~Zhang, D.~Pantazis, H.~Wang, and Q.~Li}, {\em A new
  graph gaussian embedding method for analyzing the effects of cognitive
  training}, PLoS computational biology, 16 (2020), p.~e1008186.

\bibitem{yu2018netwalk}
{\sc W.~Yu, W.~Cheng, C.~C. Aggarwal, K.~Zhang, H.~Chen, and W.~Wang}, {\em
  Netwalk: A flexible deep embedding approach for anomaly detection in dynamic
  networks}, in Proceedings of the 24th ACM SIGKDD International Conference on
  Knowledge Discovery \& Data Mining, 2018, pp.~2672--2681.

\bibitem{zachary1977information}
{\sc W.~W. Zachary}, {\em An information flow model for conflict and fission in
  small groups}, Journal of anthropological research, 33 (1977), pp.~452--473.

\bibitem{zhang2017user}
{\sc D.~Zhang, J.~Yin, X.~Zhu, and C.~Zhang}, {\em User profile preserving
  social network embedding}, in IJCAI International Joint Conference on
  Artificial Intelligence, 2017.

\bibitem{zhang2018timers}
{\sc Z.~Zhang, P.~Cui, J.~Pei, X.~Wang, and W.~Zhu}, {\em Timers: Error-bounded
  svd restart on dynamic networks}, in Thirty-Second AAAI Conference on
  Artificial Intelligence, 2018.

\bibitem{zhou2019asset}
{\sc Y.~Zhou and H.~Li}, {\em Asset diversification and systemic risk in the
  financial system}, Journal of Economic Interaction and Coordination, 14
  (2019), pp.~247--272.

\bibitem{zhu2018deep}
{\sc D.~Zhu, P.~Cui, D.~Wang, and W.~Zhu}, {\em Deep variational network
  embedding in wasserstein space}, in Proceedings of the 24th ACM SIGKDD
  International Conference on Knowledge Discovery \& Data Mining, 2018,
  pp.~2827--2836.

\bibitem{zhu2018high}
{\sc D.~Zhu, P.~Cui, Z.~Zhang, J.~Pei, and W.~Zhu}, {\em High-order proximity
  preserved embedding for dynamic networks}, IEEE Transactions on Knowledge and
  Data Engineering, 30 (2018), pp.~2134--2144.

\end{thebibliography}

\end{document}